\documentclass[runningheads]{llncs}

\usepackage{eccv}

\usepackage{eccvabbrv}

\usepackage{graphicx}
\usepackage{booktabs}

\usepackage[accsupp]{axessibility}  

\usepackage[pagebackref,breaklinks,colorlinks,citecolor=eccvblue]{hyperref}

\usepackage{orcidlink}

\usepackage{multirow}
\usepackage{booktabs}
\usepackage{makecell}
\usepackage{bbding}

\usepackage{tikz}

\usepackage{circledsteps}

\begin{document}

\title{MasterWeaver: Taming Editability and Face Identity for Personalized Text-to-Image Generation} 

\titlerunning{MasterWeaver}

\author{Yuxiang Wei$^{1, 2}$ \quad
Zhilong Ji$^{3}$ \quad
Jinfeng Bai$^{3}$ \quad
Hongzhi Zhang$^{1}$ \quad \\
Lei Zhang$^{2}$\textsuperscript{(\Envelope)} \quad
Wangmeng Zuo$^{1, 4}$\textsuperscript{(\Envelope)} \\ }

\authorrunning{Y. Wei et al.}

\institute{$^1$Harbin Institute of Technology \quad $^2$ The Hong Kong Polytechnic University  \quad $^3$Tomorrow Advancing Life \quad $^4$ Pazhou Lab Huangpu}

\maketitle

\begin{figure}[h]
   \begin{center}
   \includegraphics[width=1\linewidth]{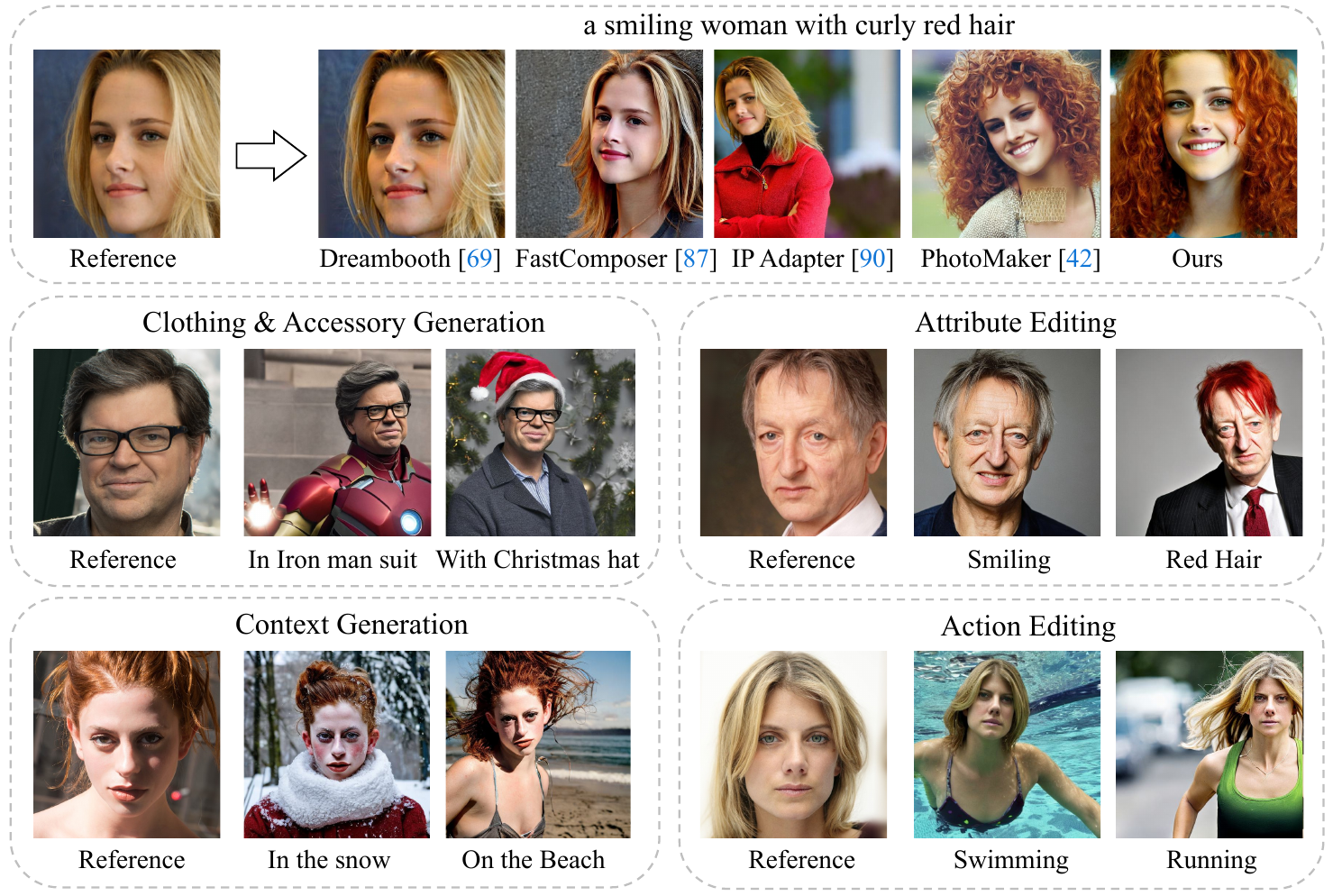}
   \end{center}
   \caption{
   With one single reference image, our MasterWeaver can generate photo-realistic personalized images with diverse clothing, accessories, facial attributes and actions in various contexts.
   In comparison with existing methods, our method exhibits superior editability while maintaining high identity fidelity. 
   } 
   \label{fig:teaser}
\end{figure}

\begin{abstract}

Text-to-image (T2I) diffusion models have shown significant success in personalized text-to-image generation, which aims to generate novel images with human identities indicated by the reference images.
Despite promising identity fidelity has been achieved by several tuning-free methods, they often suffer from overfitting issues.
The learned identity tends to entangle with irrelevant information, resulting in unsatisfied text controllability, especially on faces.
In this work, we present \textbf{MasterWeaver}, a test-time tuning-free method designed to generate personalized images with both high identity fidelity and flexible editability.
Specifically, MasterWeaver adopts an encoder to extract identity features and steers the image generation through additionally introduced cross attention.
To improve editability while maintaining identity fidelity, we propose an editing direction loss for training, which aligns the editing directions of our MasterWeaver with those of the original T2I model.
Additionally, a face-augmented dataset is constructed to facilitate disentangled identity learning, and further improve the editability.
Extensive experiments demonstrate that our MasterWeaver can not only generate personalized images with faithful identity, but also exhibit superiority in text controllability.
%
Our code can be found at \url{https://github.com/csyxwei/MasterWeaver}.


\end{abstract}

\section{Introduction}
\label{sec:intro}

Recently, text-to-image diffusion models~\cite{rombach2022high} have demonstrated superior capabilities in generating high-quality and creative images.
Building upon these advancements, personalization methods~\cite{gal2022image,ruiz2022dreambooth,wei2023elite,ye2023ip} further enable the generation of a specific visual subject (\eg, objects, animals, or people) as indicated by one or several reference images.
In this work, we focus on the personalized image generation of human identities, which has wide applications, such as personalized portrait photos~\cite{li2023photomaker}, art creation~\cite{valevski2023face0}, and visual try-on~\cite{wang2018toward}.

Earlier studies, \eg, Dreambooth~\cite{ruiz2022dreambooth} and Textual Inversion~\cite{gal2022image}, usually learned the novel identities from reference images by optimizing the word embeddings or model parameters.
Albeit the promising results, these methods require per-identity optimization, which is time-consuming and impractical for real applications.
Additionally, they often take multiple images for identity learning and suffer from poor editability in limited-data scenarios (see the top of Fig.~\ref{fig:teaser}).
Recent tuning-free methods, \eg, FastComposer~\cite{xiao2023fastcomposer} and IP Adapter~\cite{ye2023ip} trained additional visual
encoders to extract the identity information, and inject them through model tuning or adapters.
After training on human datasets, these methods could produce personalized results in several denoising steps with only one identity image as the reference.
Despite these methods having achieved promising identity fidelity, they usually suffer from overfitting issues.
Since the reference identity image and the input image used during training are from the same image, these methods tend to directly copy the reference image during generation, resulting in poor editability (see the top of Fig.~\ref{fig:teaser}).
Photomaker~\cite{li2023photomaker} took multiple images of the same identity to learn a stacked face embedding, which can be used to generate images with diverse attributes.
However, its identity fidelity is sacrificed for improving editability.

To address the above issues, we propose \textbf{MasterWeaver}, a tuning-free method for generating personalized images with both high identity fidelity and flexible text controllability.
Following~\cite{ye2023ip, wei2023elite}, MasterWeaver adopts an encoder to extract the identity features and steers the image generation through additionally introduced cross attention.
To improve the editability while keeping the identity fidelity, we propose identity-preserved editability learning.
We first propose an editing direction loss to facilitate the model training.
Specifically, we identify the editing direction in the feature space of diffusion model by inputting paired text prompts that denote an editing operation, \eg, (a photo of a person, a photo of a person with <attribute>).
As the direction's computation solely depends on the attribute difference, such a direction captures the meaningful semantic editing prior and is unrelated to identity.
By aligning the editing direction of our MasterWeaver with that of the original T2I model, we can significantly enhance the text controllability without compromising identity fidelity.

Additionally, we construct a face-augmented dataset to facilitate disentangled identity learning.
In particular, we utilize the face editing method~\cite{lyu2023deltaedit} to modify the attributes of reference identity images, which are then incorporated into our augmented dataset.
In this dataset, the reference identity image and its corresponding input image have the same identity but differ in a single attribute (\eg, hair color, style, or expression).
Such a controlled attribute misalignment can effectively facilitate model learning to extract faithful identity features disentangled with attribute details, thereby improving editability.
As illustrated in Fig.~\ref{fig:teaser}, with only one reference image, our MasterWeaver can generate photo-realistic personalized images with faithful identity and diverse contexts.

Extensive experimental evaluations demonstrate that our MasterWeaver outperforms current state-of-the-art methods. The main contributions of this work are summarized as follows:

\begin{itemize}
    \item We propose MasterWeaver, a novel method that can generate personalized images with high efficiency, faithful identity, and flexible controllability.
    \item An editing direction loss is proposed to improve the editability while keeping the identity. A face-augmented dataset is constructed to facilitate disentangled identity learning, further improving edtiability.
    \item Experimental results show that MasterWeaver can generate the target identity faithfully, while showing more flexible text controllability.
\end{itemize}

\section{Related Work}
\label{sec:related_work}

\begin{figure}[t]
   \begin{center}
   \includegraphics[width=1.0\linewidth]{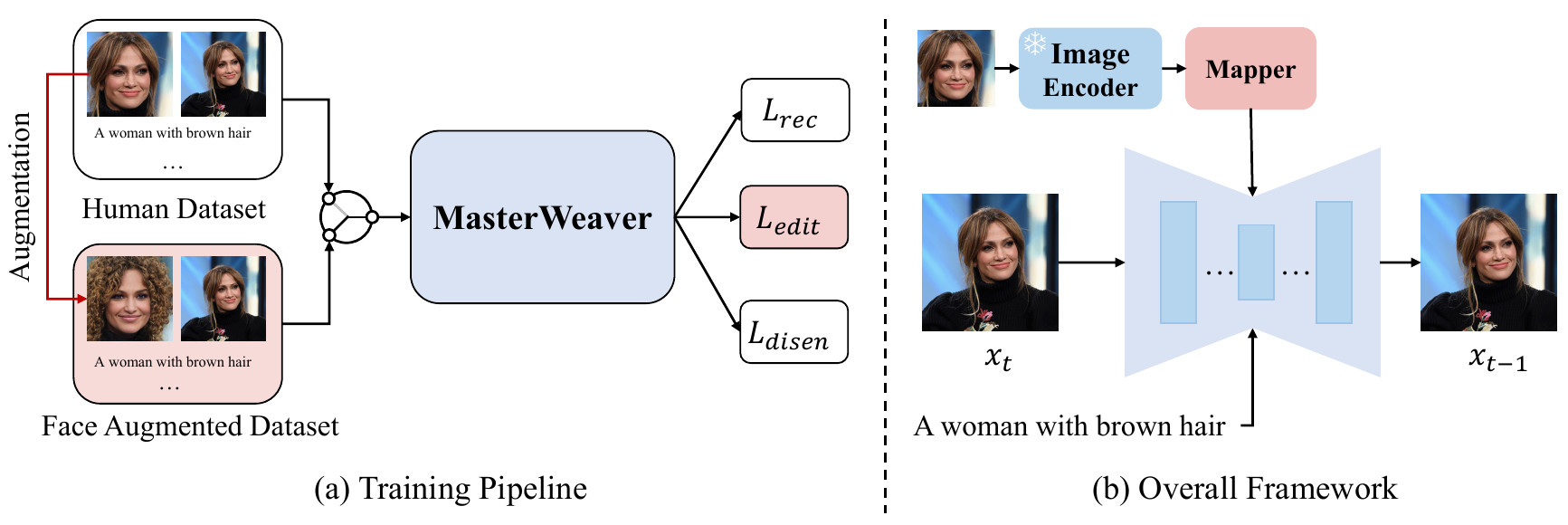}
   \end{center}
   \caption{\textbf{(a) Training pipeline of our MasterWeaver.} 
   Specifically, to improve the editability while maintaining identity fidelity, we propose an editing direction loss $\mathcal{L}_{edit}$ for training.
   Additionally, we construct a face-augmented dataset to facilitate disentangled identity learning, further improving editability.
   \textbf{(b) Framework of our MasterWeaver.} 
   It adopts an encoder to extract identity features and employ it with text to steer personalized image generation through cross attention. 
   } 
    \label{fig:method}
\end{figure}

\subsection{Text-to-Image Diffusion Models}

Recently, diffusion models~\cite{ho2020denoising, dhariwal2021diffusion} have demonstrated remarkable capabilities in generating photo-realistic images and have been widely adopted in text-to-image (T2I) generation~\cite{balaji2022ediffi,saharia2022photorealistic,ramesh2022hierarchical,ding2022cogview2,nichol2021glide,rombach2022high, betker2023improving, podell2023sdxl}.
Benefiting from the advance in language model~\cite{schuhmann2022laion, radford2019language, radford2021learning} and large-scale text-image datasets~\cite{schuhmann2022laion}, these T2I diffusion models are capable of generating textually coherent and high-quality images.
Among them, Stable Diffusion~\cite{rombach2022high} is one of the representative open-sourced models, which performs a diffusion process in the latent space to reduce computational complexity.
It has demonstrated the superior capacity in generating high-quality and diverse images, and facilitated a surge of recent advances in downstream tasks~\cite{zhang2023adding,zhang2023controlvideo,poole2022dreamfusion,lin2023improving,huang2023dreamcontrol,lv2024place,zhang2024videoelevator}.
Stable Diffusion XL (SDXL)~\cite{podell2023sdxl} employed a larger UNet and an additional text encoder to achieve superior image generation quality and enhanced textual fidelity.
Stable Diffusion is also employed as the T2I model in our experiments.

\subsection{Personalized Text-to-image Generation}

Building upon the advances in  T2I models, personalization methods~\cite{gal2022image,ruiz2022dreambooth,wei2023elite,ye2023ip} further enable the generation of specific visual concepts (\eg, objects, animals,  people) as indicated by one or several reference images.
Earlier studies~\cite{gal2022image, dong2022dreamartist, ruiz2022dreambooth, kumari2022multi, wang2023hifi, fei2023gradient, zhou2023enhancing, hua2023dreamtuner, nam2024dreammatcher, chen2023disenbooth, lee2024direct, cai2023decoupled, qiu2024controlling, liu2023cones, liu2023cones2, zhang2023compositional, hong2024comfusion, zhao2023catversion, avrahami2023break, voynov2023p+, patel2024lambda, tewel2023key, ryu2024memory, lu2024object, arar2024palp, wu2023singleinsert, hao2023vico} usually learned the novel concept from reference images by optimizing word embeddings or model parameters.
For example, Textual Inversion~\cite{gal2022image} optimized a new ``word'' embedding using a few reference images to learn the target concept.
DreamBooth~\cite{ruiz2022dreambooth} finetuned the parameters of the T2I model to align the unique identifier with the high-fidelity new concept.
To improve the computation efficiency, Custom Diffusion~\cite{kumari2022multi} only updated the key and value mapping parameters in cross attention layers.
CelebBasis~\cite{yuan2023inserting} projected the human identity into a celeb space, extracted from celebrity identities, and showed superior editability.
Though the results are promising, these methods usually take multiple images to learn the concept and require substantial time for finetuning.
To reduce the tuning time, several methods~\cite{gal2023designing, gal2023encoder} suggested initial pre-training on large datasets followed by tuning on smaller datasets to expedite the process. 
However, they still take tens of steps for finetuning, limiting their practicality.

Recent tuning-free methods~\cite{shi2023instantbooth, arar2023domain, purushwalkam2024bootpig, li2024blip, wei2023elite, zhang2023ssr, ma2023subject, chen2024subject, jia2023taming, chae2023instructbooth, hu2024instruct} train additional visual encoders to extract the concept information, and inject them via model tuning or adapters.
By training on personalization datasets, these methods could produce personalized results in tens of denoising steps with only one reference image.
For example, ELITE~\cite{wei2023elite} employed a global mapping to encode the CLIP~\cite{radford2021learning} features of concept image as textual embedding, and further improved finer details with a local mapping network.
BLIP Diffusion utilized a pre-trained BLIP2~\cite{li2023blip} as the feature extractor to produce a visual representation aligned with the text.
Additionally, several methods~\cite{wang2024instantid, ruiz2023hyperdreambooth, yan2023facestudio, valevski2023face0, yang2023diffusion, liang2024caphuman, hyung2023magicapture, chen2023photoverse, peng2023portraitbooth, wang2024stableidentity, liu2024towards, li2023stylegan, chen2023dreamidentity} are designed for personalization of human identities.
DreamIdentity~\cite{chen2023dreamidentity} crafted an editing dataset based on celebrity identity and a pre-trained stable diffusion model for editability learning.
Photomaker~\cite{li2023photomaker} collected a human dataset that consists of multiple images of the same identity and used them to learn a stacked face embedding, which could generate images with diverse attributes.

In this work, we propose the tuning-free MasterWeaver. 
Compared with existing methods, MasterWeaver could generate personalized images with high efficiency, faithful identity preservation, and flexible controllability.

\subsection{Face Editing}

Face generation and editing are popular topics in computer vision and computer graphics.
With the development of Generative Adversarial Networks (GANs)~\cite{goodfellow2014generative}, several methods have been proposed for high-quality face generation~\cite{karras2019style, karras2020analyzing} and flexible face editing~\cite{lyu2023deltaedit, shen2020interfacegan, bau2018gan, ling2021editgan, patashnik2021styleclip}.
Generally speaking, existing GAN-based face editing methods can be classified into two types.
The first type of methods~\cite{choi2018stargan, liu2019stgan, choi2020stargan} utilize image-to-image translation techniques, where the original face image is fed to the network to produce the edited image. 
The second type of methods~\cite{shen2020interfacegan, bau2018gan, ling2021editgan} first invert the given face into the latent code of pre-trained GANs~\cite{wang2022high,abdal2019image2stylegan,richardson2021encoding}, and then manipulate it with specific directions. 
Recently, with the advance of CLIP~\cite{radford2021learning}, several methods~\cite{lyu2023deltaedit, patashnik2021styleclip, gal2022stylegan} have been proposed to perform face editing based on text prompts.

In our work, we adopt the face editing method in~\cite{lyu2023deltaedit} to construct a face-augmented dataset, which can be used for disentangled identity training.

\begin{figure}[t]
   \begin{center}
   \includegraphics[width=.99\linewidth]{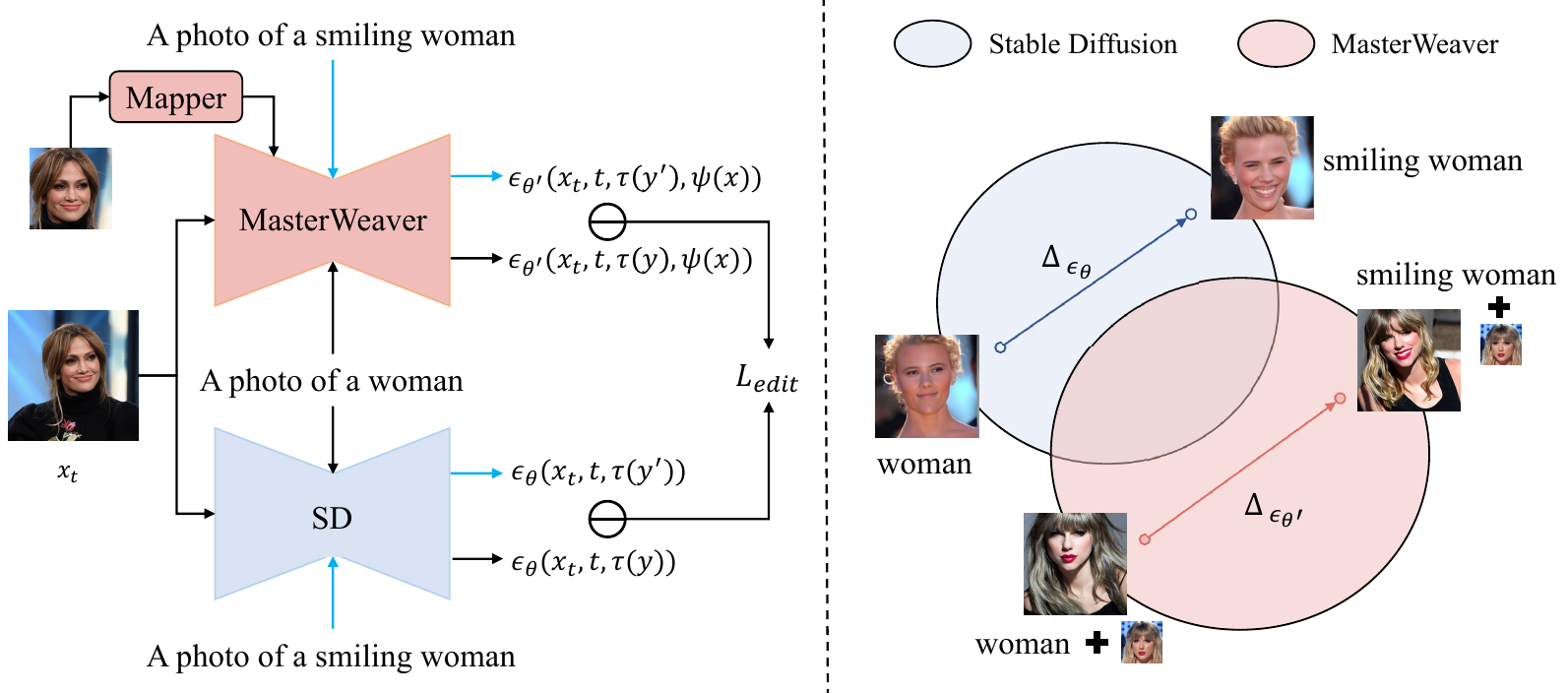}
   \end{center}
   \caption{\textbf{Illustration of Editing Direction Loss.} 
   By inputting paired text prompts that denote an editing operation, \eg, (a photo of a woman, a photo of a smiling woman), we identify the editing direction in the feature space of diffusion model.
   Then we align the editing direction of  MasterWeaver with that of original T2I model to improve the text controllability without affecting the identity.
   } 
    \label{fig:editing_loss}
\end{figure}

\section{Proposed Method}
\label{sec:method}

Given one single reference image $x$ of a specific identity, personalized T2I generation aims to generate photo-realistic images of the given identity according to the text prompts $y$. 
Nonetheless, the generated images are expected to keep a faithful identity and exhibit diversity in attributes, actions, contexts, and so on.
To achieve this goal, we propose MasterWeaver, a tuning-free method for generating personalized images with promising identity fidelity and flexible editability.
As illustrated in Fig.~\ref{fig:method}, MasterWeaver first employs an identity mapping network to encode identity feature, and incorporates it with text to guide the image generation.
To improve the model editability while keeping identity fidelity, we further propose the id-preserved editability learning, including an editing direction loss and a face-augmented dataset.
In the following, we first present an overview of the T2I model utilized in our approach (Sec.~\ref{sec:pre}).
Then, we introduce the details of the face identity injection (Sec.~\ref{sec:id_inject}) and id-preserved editability learning (Sec.~\ref{sec:edit_learning}).
Finally, we give our learning objective (Sec.~\ref{sec:loss}).

\subsection{Preliminary}
\label{sec:pre}

In this work, we employ the large-scale pre-trained Stable Diffusion (SD)~\cite{rombach2022high} as our T2I model, which can generate diverse and photo-realistic images.
It consists of two components, \ie, the autoencoder ($\mathcal{E}(\cdot)$, $\mathcal{D}(\cdot)$) and the conditional diffusion model $\epsilon_{\theta}(\cdot)$.
Specifically, the encoder $\mathcal{E}(\cdot)$ is trained to map an image $x$ to a lower dimensional latent space $z = \mathcal{E}(x)$, and the decoder $\mathcal{D}(\cdot)$ is trained to map the latent code back to the image so that $ D(\mathcal{E}(x)) \approx x $.
The conditional diffusion model $\epsilon_{\theta}(\cdot)$ is trained on the latent space of autoencoder, which can generate latent codes based on text condition $y$.
The mean-squared loss is employed to train the model:
\begin{equation}
\small
\mathcal{L}_{rec} = \mathbb{E}_{z, y, \epsilon, t}\Big[ \Vert \epsilon - \epsilon_\theta(z_{t},t, \tau(y)) \Vert_{2}^{2}\Big] \, ,
\label{eq:LDM_loss}
\end{equation}
where $\epsilon \sim \mathcal{N}(0, 1)$ denotes the unscaled noise, $t$ is the time step, $z_t$ is the latent noise at time $t$, and $\tau(\cdot)$ represents the pretrained CLIP text encoder~\cite{radford2021learning}.
During inference, it starts from a random Gaussian noise $z_T$ and iteratively denoises it to $z_0$.
The decoder then maps it to the final image $x' = \mathcal{D}(z_0)$.

Cross attention is adopted in SD to steer the generation process by text prompt.
Specifically, cross attention adopts the latent image feature $h$ and text feature $\tau(y)$ as input, and transforms them to query $Q = W_Q \cdot h$, key $K = W_K \cdot \tau (y)$ and value $V = W_V \cdot \tau(y)$ by projection layers.
$W_Q$, $W_K$, and $W_V$ are weight parameters of query, key, and value projection layers, respectively.
Then, attention maps are computed with:
\begin{equation}
\small
\text{Attention}(Q, K, V) = \text{Softmax}\left( \frac{QK^T}{\sqrt{d'}} \right) V,
\label{eq:attention}
\end{equation}
where $d'$ is the output dimension of key and query features.

\begin{figure}[t]
   \begin{center}
   \includegraphics[width=.99\linewidth]{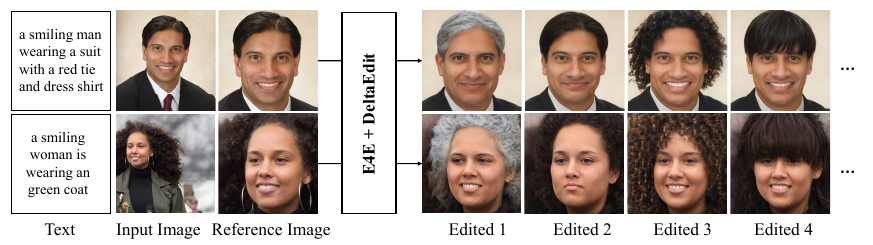}
   \end{center}
   \caption{\textbf{Construction of Face-Augmented Dataset.}
   We employ E4E~\cite{richardson2021encoding} and DeltaEdit~\cite{lyu2023deltaedit} to edit the attribute of the reference identity image, and construct the face-augmented dataset.
   } 
   \label{fig:data_pipe}
\end{figure}

\subsection{Faithful Identity Injection}
\label{sec:id_inject}

To leverage the pre-trained SD for generating the personalized images of specified identity, we first introduce a mapping network to encode the identity features, and then integrate the encoded features into the image generation process.
As shown in Fig.~\ref{fig:method} (b), we employ the pre-trained CLIP image encoder $\psi(\cdot)$ to extract the identity information.
To ensure identity fidelity, we utilize the local patch image feature from the penultimate layer of the CLIP model, which contains rich details and is sufficient to represent the target identity faithfully.
To bridge the gap between the CLIP features and the SD model, we further employ an identity (ID) mapper $M(\cdot)$, which is trained to project the CLIP features as identity feature $f$:
\begin{equation}
    f = M \circ \psi(x),
\end{equation}
where $f \in \mathbb{R}^{N \times d}$. 
$N$ denotes the number of tokens and $d$ is the feature dimension.
Here, we also mask the background of $x$ to ease the irrelevant disturbances.

To effectively integrate the identity feature $f$ into the generation, we adopt the dual cross attention mechanism~\cite{ye2023ip, wei2023elite}: $\text{Attention} (Q, K^f, V^f)$, where $K^f = W^f_K \cdot f$ and $V^f = W^f_V \cdot f$ represent the projected key and value matrices of the identity information, respectively.
$W^f_K$ and $W^f_V$ are two additionally introduced projection layers in each cross attention block of SD, which are optimized with our ID mapper simultaneously.
To fuse the identity information with the text information, we sum them by:
\begin{equation}
\small
\textit{Out} = \text{Attention} (Q, K, V) + \lambda \text{Attention} (Q, K^f, V^f), 
\label{eq:fuse}
\end{equation}
where $\lambda$ is a trade-off hyperparameter and set as $1$ during training.

\begin{figure}[t]
   \begin{center}
   \includegraphics[width=.99\linewidth]{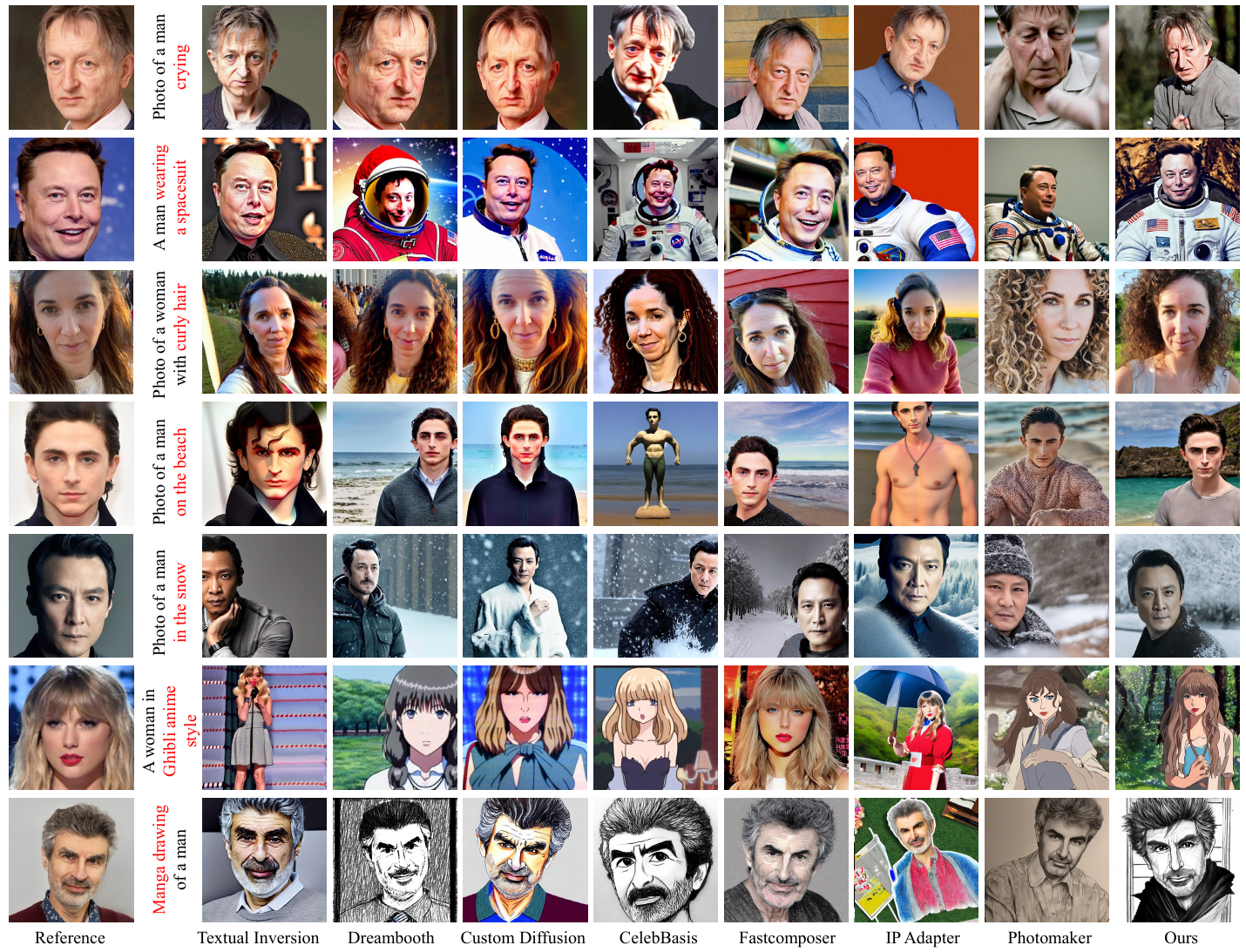}
   \end{center}
   \caption{\textbf{Visual comparison of different methods.} 
   All images are generated using the single reference image shown on the left. 
   Our MasterWeaver can generate high-quality images with flexible editability and faithful identity.
   Zoom in for a better view.
   } 
    \label{fig:compare_single}
\end{figure}

\subsection{ID-Preserved Editability Learning}
\label{sec:edit_learning}

Although the proposed method in Sec.~\ref{sec:id_inject} can generate personalized images with faithful identity, its text controllability is limited.
Since the reference identity image and input image used in training are from the same image (as shown in Fig.~\ref{fig:data_pipe} left), when we train the model under the reconstruction paradigm, the learned identity feature is inevitably entangled with facial attributes (\eg, hair, pose and expression).
Such an entanglement weakens the control of the text over generated images, and the model tends to directly copy reference images during generation, resulting in poor editability and diversity.
To improve editability while keeping a faithful identity, we propose id-preserved editability learning, which consists of an editing direction loss and a face-augmented dataset.

\noindent \textbf{Editing Direction Loss.}
To improve the text controllability, we first propose an editing direction loss to facilitate the model learning.
The SD model is well-recognized for its superior text controllability and can generate images that closely align with the provided textual descriptions.
Intuitively, we can employ its editing capability to regularize MasterWeaver.
As shown in Fig.~\ref{fig:editing_loss}, by leveraging paired text prompts that indicate an editing operation, \eg, ($y = $ a photo of a person, $y' = $ a photo of a person with <attribute>), we can identify an editing direction in the feature space of diffusion UNet model:
\begin{align}
    \Delta_{\epsilon_{\theta}}(y, y') = \epsilon_{\theta}(z_t, t, \tau(y')) - \epsilon_{\theta}(z_t, t, \tau(y)).
\end{align}
Such an editing direction captures the meaningful prior of SD for semantic editing.
Then, we align the editing directions of MasterWeaver with those of SD during training:
\begin{equation}
    \mathcal{L}_{edit} = 1 - \text{Sim} ( m \cdot \Delta_{\epsilon_{\theta}}(y, y') , m \cdot \Delta_{\epsilon_{\theta'}}(y, y', x) ),
\end{equation}
where $\epsilon_{\theta'}$ denotes our MasterWeaver model and $\text{Sim}(\cdot, \cdot)$ denotes the calculation of cosine similarity.
$m$ is the mask of the facial region to ensure the loss is only calculated on the facial area.
Obviously, if MasterWeaver neglects the text prompts during generation, its editing directions will be different from those of SD, resulting in a large loss. 
Besides, the calculation of editing direction is solely based on the text difference, and the encoded semantic information is unrelated to the identity.
Therefore, training with our proposed direction loss can improve the model's editability significantly, with moderate sacrifice of identity fidelity.

Furthermore, our editing direction $\Delta_{\epsilon_{\theta}}(y, y')$ can be seen as a variant of DDS~\cite{hertz2023delta}, 
which is designed to optimize the edited image by minimizing the delta noise: $\mathcal{L}_{dds} = \Vert \epsilon_{\theta}(z_t, t, \tau(y)) - \epsilon_{\theta}(\hat{z}_t, t, \tau(y')) \Vert_2^2$, where $z$ and $\hat{z}$ are original and edited images, $y$ and $y'$ are source and target prompts.
In $\mathcal{L}_{edit}$, editing direction $\Delta_{\epsilon_{\theta}}(y, y') = \epsilon_{\theta}(z_t, t, \tau(y')) - \epsilon_{\theta}(z_t, t, \tau(y))$ is used to represent the editing prior of original model.
By aligning the editing direction with that of the original model, MasterWeaver can inherit its editability.
In our implementation, we use the same $z_t$ for both source prompt $y$ and target prompt $y'$, and the direction is calculated in feature space instead of noise space.
We have collected several attribute-related prompt pairs for training, including hair, age, expression, \etc.
More details can be found in the Suppl.

\noindent \textbf{Face-Augmented Dataset Construction.}
As we analyzed, one of the reasons behind the insufficient editability is that the learned model entangles the identity with irrelevant attributes and tends to copy the reference image directly, neglecting the text prompt. 
To address this issue, we build a face-augmented dataset designed for disentangled identity training.
As illustrated in Fig.~\ref{fig:data_pipe}, we employ the E4E~\cite{richardson2021encoding} and DeltaEdit~\cite{lyu2023deltaedit} to perform the attribute editing for the reference identity image.
Then, we construct our face-augmented dataset by combining the input image, text prompt, and edited face images.
In this dataset, the reference identity image and corresponding input image have the same identity but differ in one specific attribute (\eg, hair color, style, or expression).
Such a controlled attribute misalignment can effectively facilitate model learning to extract faithful identity features disentangled with attribute details, thereby improving editability.
To maintain identity fidelity, we filter to exclude edited face images with lower face similarity to reference images.
In our experiments, we utilize a mix of face-augmented and original datasets for balanced editability and identity fidelity.
More details of the dataset are provided in the Suppl.

\subsection{Learning Objective}
\label{sec:loss}

Following~\cite{li2023stylegan}, we employ a background disentanglement loss to encourage the identity feature only controlling the generation of facial region.
Specifically, when the identity feature changes, we regularize the background to be unchanged:
\begin{equation}
     \mathcal{L}_\textit{disen}= \Vert (1 - m) \cdot [ \epsilon_{\theta'}(z_t, t, \tau(y), f) -  \epsilon_{\theta'}(z_t, t, \tau(y), A(f)) ] \Vert ,
    \label{eqn:reg}
\end{equation}
where $f = M \circ \psi(x)$ is the extracted identity feature and $1 - m$ denotes the mask of background region.
$A(\cdot)$ denotes the augmentation operation.

The overall learning objective is defined as:
\begin{equation}
    \small
    \setlength{\abovedisplayskip}{5pt}
    \setlength{\belowdisplayskip}{5pt}
    \mathcal{L}= \mathcal{L}_\textit{rec} + \lambda_{edit} \cdot \mathcal{L}_{edit} +\lambda_{disen} \cdot \mathcal{L}_{disen},
    \label{eqn:objetive}
\end{equation}
where $\lambda_{disen}$ and $\lambda_{edit}$ denote the trade-off parameters.

\begin{figure}[t]
   \begin{center}
   \includegraphics[width=.99\linewidth]{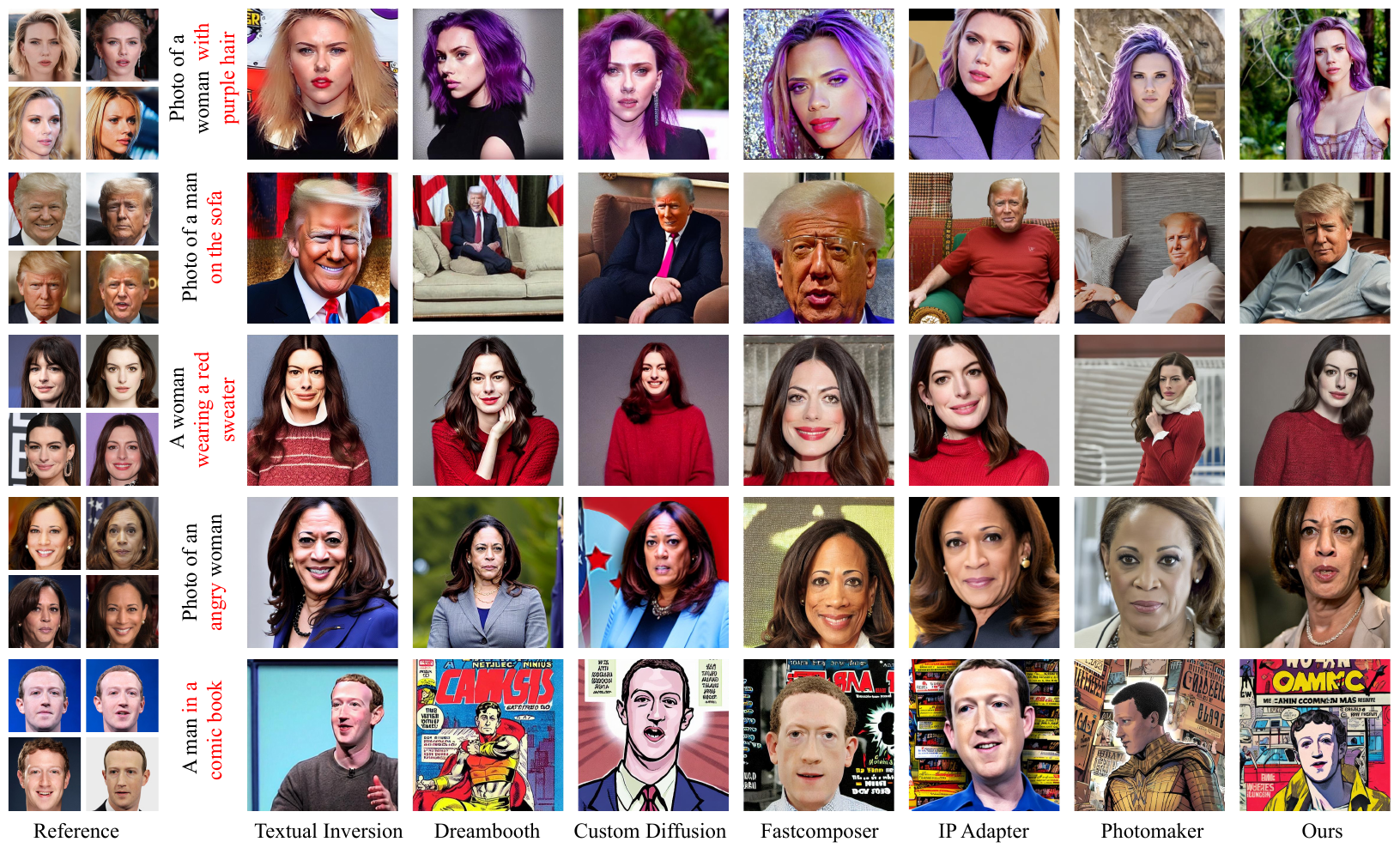}
   \end{center}
   \caption{\textbf{Visual comparison of different methods.}
   All images are generated using the four reference images shown on the left. 
   Our MasterWeaver consistently shows better editability and identity.
   Zoom in for a better view.
   } 
   \label{fig:compare_multi}
\end{figure}

\begin{table}[t]
    \centering
    \caption{ \textbf{Quantitative comparison under single reference setting}. 
    We employ CLIP-T metric to measure the text alignment, and utilize DINO, CLIP-I and Face Similarity metrics to measure the identity fidelity. The best result is shown in \textbf{bold}, and the second best is \underline{underlined}. }
    \label{tab:quantitative_single}
    \resizebox{1\linewidth}{!}{
    \setlength{\tabcolsep}{1.5mm}
    \begin{tabular}{lccccc} 
        \toprule
         &  CLIP-T ($\uparrow$) &  CLIP-I ($\uparrow$) &  DINO ($\uparrow$) &  Face Sim. ($\uparrow$) &  Speed (s, $\downarrow$) \\ 
        \midrule
         Textual Inversion~\cite{gal2022image} &  0.167&  0.669&  0.556&  0.423&    3000  \\
         DreamBooth~\cite{ruiz2022dreambooth} &  0.223&  0.641&   0.509&  0.469&   908 \\ 
         Custom Diffusion~\cite{kumari2022multi} &  0.210&  \underline{0.726}&  \textbf{0.647}&  \underline{0.626}&      548 \\ 
         CelebBasis~\cite{yuan2023inserting} &  0.214&  0.651&  0.501&  0.439&   490 \\ 
         FastComposer~\cite{xiao2023fastcomposer}&  0.201&  \textbf{0.747}&  \underline{0.641}&  \textbf{0.631}&   \textbf{3} \\ 
         IPAdapter~\cite{ye2023ip}&  0.192&  0.714&  0.632&  0.607&   \textbf{3} \\ 
         PhotoMaker~\cite{li2023photomaker} &  \underline{0.231} &  0.667&  0.497 &  0.544 &  10 \\ 
         Ours &  \textbf{0.232}&  \underline{0.726}&  0.638 & \textbf{0.631}&   \underline{4}\\ 
        \bottomrule
    \end{tabular}
    }
\end{table}

\section{Experiments}
\label{sec:exp}

\subsection{Experimental Settings}

\noindent \textbf{Training dataset.}
To train our MasterWeaver, we build a dataset including about 160k text-image pairs from the LAION-Face dataset~\cite{zheng2022general}.
We have also prepared the corresponding captions and face masks for model training.
The detailed pipeline of dataset is provided in the Suppl.
Moreover, as described in Sec.~\ref{sec:edit_learning}, we have built a face-augmented dataset based on filtered images.

\noindent \textbf{Evaluation dataset.}
To evaluate the one-shot personalization, we randomly select 300 identities from the CelebA-HQ dataset~\cite{CelebAMask-HQ}, and each identity has one image as the reference.
Besides, our method can be extended for personalized generation using multiple reference images without additional training.
Here, we also evaluate our method with multiple images as the reference.
Following~\cite{li2023photomaker}, we collect a dataset consisting of 25 identities, and each identity contains four images for evaluation.
For quantitative analysis, we employ 50 prompts encompassing a range of clothing, styles, attributes, actions, and backgrounds.
We randomly generate five images for each identity-prompt pair.
More details are provided in Suppl.
For visual comparison, we utilize the identity images collected from existing methods for a convenient comparison.
Unless specifically mentioned, all images presented in this paper are produced using a single reference image.

\noindent \textbf{Implementation details.}
We employ SD V1.5 in our experiments, and our model is trained using a batch size of 16 and a learning rate of 1e-6.
To calculate the $\mathcal{L}_{edit}$, we utilize the output features of the cross attention blocks in the first decoder layer.
$\lambda_{edit}$ is set as 0.01 and $\lambda_{disen}$ is set as 1. 
To enable classifier-free guidance, we use a probability of 0.05 to drop text and image individually, and a probability of 0.05 to drop text and image simultaneously.
All experiments are conducted on 4$\times$A800 GPUs with AdamW~\cite{loshchilov2017decoupled} optimizer.
During image generation, we use 50 steps of the LMS sampler, and set the scale of classifier-free guidance as 5. 
More details of our model and implementation are provided in the Suppl.

\begin{table}[t]
    \centering
    \caption{ \textbf{Quantitative comparisons under multiple references (\ie, 4 images) setting}. The best result is shown in \textbf{bold}, and the second best is \underline{underlined}. }
    \label{tab:quantitative_multi}
    \resizebox{1\linewidth}{!}{
    \setlength{\tabcolsep}{1mm}
    \begin{tabular}{lccccc} 
        \toprule
         &  CLIP-T ($\uparrow$) &  CLIP-I ($\uparrow$) &  DINO ($\uparrow$) &  Face Sim. ($\uparrow$) &  Speed (s, $\downarrow$) \\ 
        \midrule
         Textual Inversion~\cite{gal2022image} &  0.172&  0.729&  0.594&  0.618&   3000 \\ 
         DreamBooth~\cite{ruiz2022dreambooth} &  0.220&  0.704&   0.511&  0.524&   908\\ 
         Custom Diffusion~\cite{kumari2022multi} &  0.218&  \underline{0.754}&  \textbf{0.598}&  0.615&    548 \\ 
          FastComposer~\cite{xiao2023fastcomposer}&  0.210 &  0.720 &  0.582&  0.622&     \textbf{3}\\ 
         IPAdapter~\cite{ye2023ip}&  0.200&  \textbf{0.757}&  0.592&  \underline{0.632}&   \textbf{3}\\ 
         PhotoMaker~\cite{li2023photomaker} &  \underline{0.222}&  0.663&  0.520 &  0.547 & 10 \\ 
         Ours &  \textbf{0.230}&  0.752&  \underline{0.596} & \textbf{0.646} &    \underline{4}\\ 
        \bottomrule
    \end{tabular}
    }
\end{table}

\noindent \textbf{Evaluation metrics.}
Following Dreambooth~\cite{ruiz2022dreambooth}, we employ CLIP-T metric to measure the text alignment of generated images, and utilize DINO and CLIP-I metrics to measure the identity fidelity.
We also employ the FaceNet~\cite{schroff2015facenet} to compute the face similarity in the detected facial regions as another measure of identity fidelity.
Moreover, we adopt the inference speed to evaluate the efficiency of each method.
For those optimization-based methods, we combine the time required for optimization with the inference time as overall speed. 
All the inference speeds are tested on a single NVIDIA Tesla V100 GPU.

\begin{figure}[t]
   \begin{center}
   \includegraphics[width=1\linewidth]{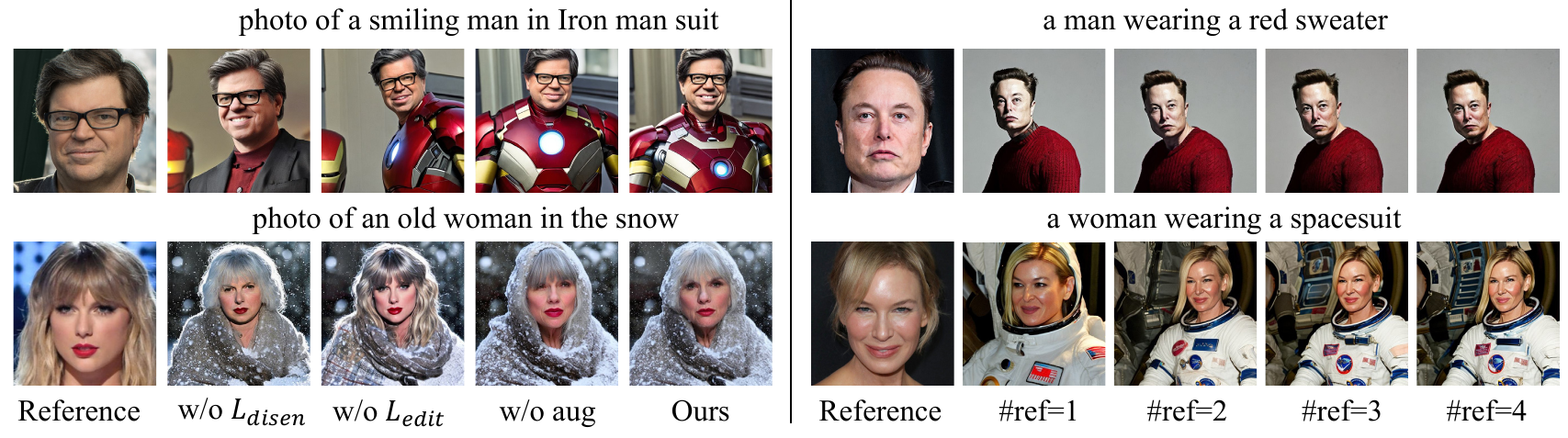}
   \end{center}
   \caption{
   \textbf{Ablation Study.} The proposed losses and augmented dataset (denotes as aug) significantly improve the editability of the model.
   Besides, the identity fidelity consistently improves as the number increases.  } 
   \label{fig:ablation}
\end{figure}

\subsection{Qualitative Comparison}

To demonstrate the effectiveness of our MasterWeaver, we conduct the comparisons with existing methods, including Textual Inversion~\cite{gal2022image}, Dreambooth~\cite{ruiz2022dreambooth}, Custom Diffusion~\cite{kumari2022multi}, CelebBasis~\cite{yuan2023inserting}, Fastcomposer~\cite{xiao2023fastcomposer}, IP Adapter~\cite{ye2023ip}, and Photomaker~\cite{li2023photomaker}.
For a fair comparison, all methods utilize the SD 1.5 version, except for Photomaker, which uses the SDXL model.

Fig.~\ref{fig:compare_single} presents visual comparisons with existing methods using a single image as a reference.
As depicted, MasterWeaver can generate photo-realistic personalized images in diverse scenarios, including modifications to attributes, clothing, background, and style.
Optimization-based methods, \ie, Textual Inversion, Dreambooth, and Custom Diffusion, suffer from the overfitting in this limited-data scenario, leading to limited text alignment and compromised identity fidelity (\eg, rows 1$\sim$3).
Tuning-free methods, \ie, Fastcomposer and IP Adapter maintain high identity fidelity but fall short in editing style and facial attributes (\eg, face attributes in rows 1 and 3, and the style in the last two rows). 
Photomaker shows improved text controllability; however, it sometimes fails to generate faithful identity (\eg, rows 1, 2, and 5).
In comparison, our MasterWeaver can generate personalized images that are faithful to both reference identity and text prompts.
Fig.~\ref{fig:teaser} illustrates more results generated by our method. MasterWeaver can generate photo-realistic images with diverse clothing, accessories, facial attributes, and actions.
Additionally, it allows for simultaneously editing multiple attributes.

MasterWeaver can also be directly extended for personalized image generation with multiple reference images of the same identity.
Specifically, during inference, we simply concatenate the identity features of different images along the token dimension.
As shown in Fig.~\ref{fig:ablation}, using more reference images improves the identity fidelity of the generated personalized images.
Furthermore, we compare our method with competing methods in the setting with multiple reference images (\ie, with four reference images).
From Fig.~\ref{fig:compare_multi}, one can see that MasterWeaver still outperforms competitors in identity fidelity and text controllability.
More qualitative results can be found in Suppl.

\begin{table}[t]
    \centering
    \caption{\textbf{Ablation Study.} The proposed background disentanglement loss $L_{disen}$, editing direction loss $L_{edit}$, and the face-augmented dataset (denote as aug) can improve the editability of MasterWeaver, while keeping the identity fidelity.}
    \label{tab:ablation}
    \resizebox{0.8\linewidth}{!}{
    \setlength{\tabcolsep}{1.4mm}
    \begin{tabular}{lcccc} 
        \toprule
         &  CLIP-T ($\uparrow$) &  CLIP-I ($\uparrow$) &  DINO ($\uparrow$) &  Face Sim. ($\uparrow$) \\ 
        \midrule
         Ours w/o $\mathcal{L}_{disen}$  &  0.226& 0.724&  0.633&  0.617\\ 
         Ours w/o $\mathcal{L}_{edit}$ &  0.209&  \textbf{0.739}&  \textbf{0.643}&  \textbf{0.637} \\ 
         Ours w/o aug &  0.221&  0.730&  0.640&  0.634 \\ 
         Ours &  \textbf{0.232}&  0.726&  0.638 & 0.631 \\
        \bottomrule
    \end{tabular}
    }
\end{table}

\subsection{Quantitative Comparison}

In addition to the qualitative comparisons, we further conduct the quantitative evaluation to validate the performance of MasterWeaver. 
Table~\ref{tab:quantitative_single} reports the comparisons conducted with a single reference image.
From the table, we see that MasterWeaver outperforms existing state-of-the-art methods regarding text alignment and face similarity, demonstrating its ability to generate personalized images that maintain consistent identity and good alignment with texts.
Moreover, MasterWeaver exhibits a competitive inference speed, requiring only 4s to generate an image on a single V100 GPU.
Table~\ref{tab:quantitative_multi} reports similar findings in a scenario utilizing four reference images, further demonstrating the effectiveness of  MasterWeaver.
While MasterWeaver's scores for CLIP-I and DINO are not the highest, it should be noted that these metrics are not specifically designed for facial analysis and can be affected by extraneous factors such as pose.
Additionally, we have conducted a user study to compare MasterWeaver with other methods. 
The detailed results are provided in Suppl.

\subsection{Ablation Study}

We have conducted ablation studies to evaluate the roles of various components in our method, including the background disentanglement loss $\mathcal{L}_{disen}$, editing direction loss $\mathcal{L}_{edit}$, and the face-augmented dataset.

\noindent \textbf{Effect of the disentanglement loss $\mathcal{L}_{disen}$.}
We first evaluate the impact of $\mathcal{L}_{disen}$.
As illustrated in Fig.~\ref{fig:ablation}, without $\mathcal{L}_{disen}$, the model fails to generate images with desired clothing.
This suggests that $\mathcal{L}_{disen}$ could ease the influence of identity information on the background, thereby enhancing the editability of the generated images.
From Table~\ref{tab:ablation}, both text alignment and face similarity metrics decrease without $\mathcal{L}_{disen}$, demonstrating its effectiveness.

\noindent \textbf{Effect of the editing direction loss $\mathcal{L}_{edit}$.}
We then assess the effect of $\mathcal{L}_{edit}$.
As shown in Fig.~\ref{fig:ablation}, without  $\mathcal{L}_{edit}$, MasteWeaver fails to generate the identities with proper facial attributes (\ie, old woman).
From Table~\ref{tab:ablation}, we see that the CLIP-T score drops significantly without $\mathcal{L}_{edit}$.
Though the preservation of identity drops slightly, it is acceptable.

\noindent \textbf{Effect of the face-augmented dataset.}
We further evaluate the effect of our constructed face-augmented dataset.
Fig.~\ref{fig:ablation} demonstrates that the model trained with the face-augmented dataset exhibits improved editability (\eg, with the augmented dataset, MasterWeaver successfully generates the personalized image with a smiling expression).
The results in Table~\ref{tab:ablation} further confirm its effectiveness, as the text alignment score increases when using the augmented dataset. More ablations are provided in Suppl.

\section{Conclusion}
\label{sec:conclusion}

In this work, we proposed MasterWeaver, a novel tuning-free method capable of generating personalized images with high efficiency, faithful identity, and flexible editability. 
The proposed editing direction loss and face-augmented dataset significantly improved the model's editability while maintaining identity fidelity.
Extensive experiments have demonstrated that MasterWeaver outperformed the state-of-the-art methods and generated photo-realistic images that are faithful to both identity and texts. 
Our method was versatile for various applications, including personalized digital content creation and artistic endeavors.
Moreover, the proposed editing direction loss had the potential to be applied to other domains (\eg, animals and objects), enlarging its applicability.

\noindent \textbf{Acknowledgement.} This work was supported in part by the National Key R\&D Program of China (2021YFF0900500), and the National Natural Science Foundation of China (NSFC) under Grant No. 62441202, and the Hong Kong RGC RIF grant (R5001-18).

%
%
\bibliographystyle{splncs04}
\bibliography{arxiv}

\renewcommand{\thefigure}{S\arabic{figure}}
\renewcommand{\thetable}{S\arabic{table}}
\renewcommand{\thesection}{S\arabic{section}}
\renewcommand{\theequation}{S\arabic{equation}}

\title{MasterWeaver: Taming Editability and Face Identity for Personalized Text-to-Image Generation \\
-- Supplementary Materials -- } 

\titlerunning{MasterWeaver}

\author{Yuxiang Wei$^{1, 2}$ \quad
Zhilong Ji$^{3}$ \quad
Jinfeng Bai$^{3}$ \quad
Hongzhi Zhang$^{1}$ \quad \\
Lei Zhang$^{2}$\textsuperscript{(\Envelope)} \quad
Wangmeng Zuo$^{1, 4}$\textsuperscript{(\Envelope)} \\ }


\authorrunning{Y. Wei et al.}

\institute{$^1$Harbin Institute of Technology \quad $^2$ The Hong Kong Polytechnic University  \quad $^3$Tomorrow Advancing Life \quad $^4$ Pazhou Lab Huangpu}

\maketitle

The following materials are provided in this supplementary file:
\begin{itemize}
    \setlength{\itemsep}{2pt}
    \setlength{\parsep}{0pt}
    \setlength{\parskip}{0pt}
    \item Sec.~\ref{sec:more_implementation}: more implementation details, including the detailed framework, the editing direction loss, and the training details.
    \item Sec.~\ref{sec:more_dataset}: more dataset details, including the training dataset, face-augmented dataset, and evaluation dataset.
    \item Sec.~\ref{sec:more_ablation}: more ablation studies, including the layer selection of $\mathcal{L}_{edit}$, the value of $\lambda$, and the number of reference images.
    \item Sec.~\ref{sec:more_comparison}: more comparisons and results.
    \item Sec.~\ref{sec:limitation}: limitation and social impact.
\end{itemize}

\section{More Implementation Details}
\label{sec:more_implementation}

\begin{figure}[t]
   \begin{center}
   \includegraphics[width=.89\linewidth]{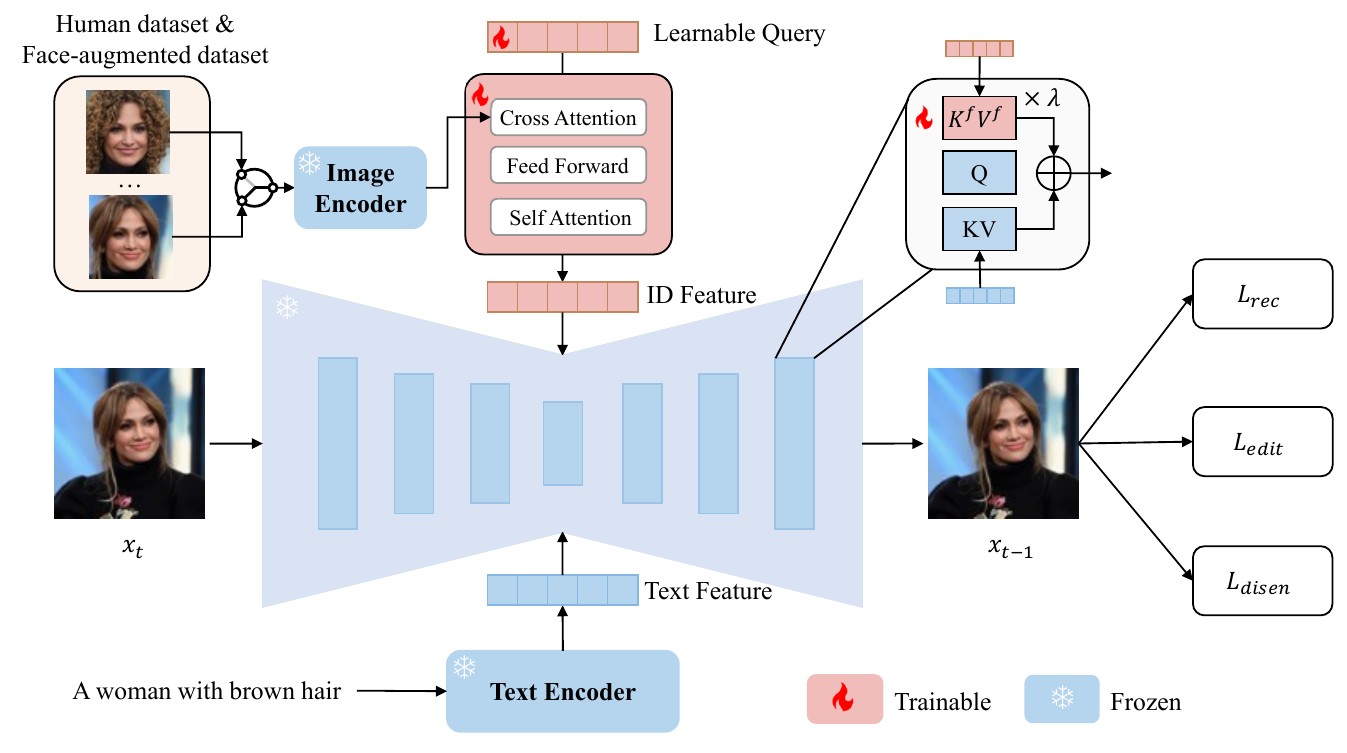}
   \end{center}
   \caption{\textbf{Framework of our MasterWeaver.}
   } 
   \label{fig:framework}
\end{figure}

\subsection{Framework of MasterWeaver}

Fig.~\ref{fig:framework} illustrates the detailed framework of MasterWeaver.
As shown in the figure, the ID Mapper consists of a stack of cross attention, feed forward layer and self attention layer.
To effectively incorporate the identity feature $f$ with text feature to steer the personalized generation, we adopt the dual cross attention mechanism~\cite{ye2023ip, wei2023elite}.
As illustrated in Fig.~\ref{fig:framework}, for each cross attention block of SD, we further introduce two learnable projection layers $W^f_K$ and $W^f_V$.
The identity feature is then integrated by $\text{Attention} (Q, K^f, V^f)$, where $K^f = W^f_K \cdot f$ and $V^f = W^f_V \cdot f$ represent the projected key and value matrices of the identity information, respectively.
To fuse the identity information with the text information, we sum them by:
\begin{equation}
\small
\textit{Out} = \text{Attention} (Q, K, V) + \lambda \text{Attention} (Q, K^f, V^f), 
\label{eq:fuse}
\end{equation}
where $\lambda$ is a trade-off hyperparameter and set as $1$ during training.
During training, we optimize the parameters of ID Mapper and the projection layers simultaneously, and keep the parameters of SD fixed.

\subsection{Editing Direction Loss}

To calculate the editing direction loss $\mathcal{L}_{edit}$, we utilize the output features from the cross attention blocks in the first decoder layer.
Specifically, given the source prompt $y$, target prompt $y'$ and reference image $x$, the editing directions of the $l$-th cross attention block for SD and MasterWeaver are $\Delta^l_{\epsilon_{\theta}}(y, y') \in \mathbb{R}^{H \times W \times D}$ and $\Delta^l_{\epsilon_{\theta'}}(y, y', x) \in \mathbb{R}^{H \times W \times D}$, where $H$, $W$, and $D$ represent the height, width, and dimension of the output feature, respectively.
Then, the editing direction loss is calculated as follows:
\begin{equation}
    \mathcal{L}_{edit} =  \sum_{l} \mathcal{L}^l_{edit},
\end{equation}
\begin{equation}
    \mathcal{L}^l_{edit} =  \frac{1}{H \times W} \sum_{h=1}^{H} \sum_{w=1}^{W} \left[ 1 - \text{Sim} [ (m \cdot \Delta^l_{\epsilon_{\theta}})_{h, w}, (m \cdot \Delta^l_{\epsilon_{\theta'}})_{h, w} ] \right].
\end{equation}
For conciseness, we have omitted the $y$, $y'$ and $x$.
The source and target prompts employed in our experiments are listed in Table~\ref{tab:editing_prompts}.
For each training iteration, we randomly select a source and target prompt pair to compute the loss.

\subsection{Training Details}

\noindent \textbf{Textual Inversion}~\cite{gal2022image}.
We use the official implementation of Textual Inversion\footnote{https://github.com/rinongal/textual\_inversion}, training it using SD V1.5.
For each identity, the experiment is conducted with a batch size of 1 and a learning rate of 0.005 for 5,000 steps. 
The new token is initialized with the word ``person''. 

\noindent \textbf{Custom Diffusion}~\cite{kumari2022multi}.
We use the official implementation of Custom Diffusion\footnote{https://github.com/adobe-research/custom-diffusion}, and train it with SD V1.5. 
During training, the batch size is set to 1 and the learning rate to 1e-5.
We generate 400 regularization images using SD V1.5 with 50 steps of DDIM sampling, prompted by the text ``\texttt{A photo of a person}''.
The model is trained for 300 steps.

\noindent \textbf{DreamBooth}~\cite{ruiz2022dreambooth}.
We use the third-party implementation of DreamBooth\footnote{https://github.com/XavierXiao/Dreambooth-Stable-Diffusion}, and train it with SD V1.5.
Training is done by finetuning both the U-net diffusion model and the text transformer. 
The training batch size is 1 and the learning rate is set as 1e-6. 
The regularization images are generated with 50 steps of the DDIM sampler with the text prompt ``\texttt{A photo of a person}''.
For each identity, we train the model for 800 steps.

\noindent \textbf{CelebBasis}~\cite{yuan2023inserting}.
We adopt the official implementation of CelebBasis\footnote{https://github.com/ygtxr1997/CelebBasis}, and train it with SD V1.5. 
Training is conducted with a batch size of 2 for 800 steps.
The learning rate is set to 0.005.

\noindent \textbf{FastComposer}~\cite{xiao2023fastcomposer}.
We employ the official implementation and pre-trained models from FastComposer\footnote{https://github.com/mit-han-lab/fastcomposer}, which is based on stable diffusion v1.5.

\noindent \textbf{IP Adapter}~\cite{ye2023ip}.
We utilize the official implementation of IP Adapter, and employ its face version for comparison\footnote{https://github.com/tencent-ailab/IP-Adapter/blob/main/ip\_adapter-plus-face\_demo.ipynb}. 

\noindent \textbf{Photomaker}~\cite{li2023photomaker}.
We use the official implementation and pre-trained models of Photomaker\footnote{https://github.com/TencentARC/PhotoMaker}, which is trained based on SDXL. 

\noindent \textbf{Our MasterWeaver.}
We employ SD V1.5 in our experiments, and our model is trained using a batch size of 16 and a learning rate of 1e-6.
We initially train our model without $\mathcal{L}_{edit}$ and the face-augmented dataset for 100k iterations.
Subsequently, we finetune it with $\mathcal{L}_{edit}$ and the face augmented dataset for an additional 20k iterations.
$\lambda_{edit}$ is set as 0.01 and $\lambda_{disen}$ is set as 1. 
To enable classifier-free guidance, we use a probability of 0.05 to drop text and image individually, and a probability of 0.05 to drop text and image simultaneously.
All experiments are conducted on 4$\times$A800 GPUs with AdamW~\cite{loshchilov2017decoupled} optimizer.

\section{More Dataset Details}
\label{sec:more_dataset}

\subsection{Training Dataset}
To train our MasterWeaver, we first build a high-quality human dataset including approximately 160k text-image pairs from the LAION-Face dataset~\cite{zheng2022general}.
Specifically, we first utilize the dlib tool\footnote{https://github.com/davisking/dlib} to detect face landmarks and filter out the images with small faces (less than 200 $\times$ 200). 
Then, we perform square cropping on the images based on the detected facial landmarks, ensuring that the facial region would occupy more than 4\% of the image post-cropping.
To prepare the text caption, we employ BLIP2~\cite{li2023blip} to generate ten captions for each cropped image and select the caption with the highest CLIP score as the final choice.
The facial masks used in Eqns.~{\color{red} 6} and {\color{red} 7} are generated based on the extracted landmarks.
The reference identity image is aligned following the FFHQ~\cite{karras2019style}.
To ease the impact of background, the background of the reference image is masked using the mask extracted by a pre-trained face parsing model\footnote{https://github.com/zllrunning/face-parsing.PyTorch}.

\subsection{Face-Augmented Dataset}

To construct our face-augmented dataset,  we employ the E4E~\cite{richardson2021encoding} and DeltaEdit~\cite{lyu2023deltaedit} to perform the attribute editing for the reference identity image.
Table~\ref{tab:augment_prompts} lists the prompts we used for editing, which cover various attributes, \eg, hair and expression, \etc. 
After filtering images with low face similarity, we finally obtain $\sim$90k augmented images for training.

\subsection{Evaluation Dataset}

To evaluate the one-shot personalization, we randomly select 300 identities from the CelebA-HQ dataset~\cite{CelebAMask-HQ}, and each identity has one image as the reference.
For quantitative analysis, we employ 50 prompts that encompass a range of clothing, styles, attributes, actions, and backgrounds.
The full prompts are listed in Table~\ref{tab:evaluate_prompts}.
Following~\cite{li2023photomaker}, we collect a dataset consisting of 25 identities listed in Table~\ref{tab:evaluate_ids}, and each identity contains four images for evaluation under the multi-reference image setting.
During evaluation, the identity embedding for FastComposer is computed as the mean embedding of the four images. 
For IP Adapter, the identity feature is constructed by concatenating the features from all the four images, similar to our approach.

\begin{table}[t]
    \centering
    \caption{\textbf{ID names used for multi-reference evaluation.} For each name, we collect four images totally.}
    \begin{tabular}{llll}
    \toprule
    \multicolumn{4}{c}{Evaluation IDs}\\
    \midrule
    \Circled{1} Alan Turing &\Circled{2} Albert Einstein &\Circled{3}  Anne Hathaway &\Circled{4} Audrey Hepburn \\
    \Circled{5} Barack Obama &\Circled{6} Bill Gates &\Circled{7} Donald Trump &\Circled{8} Dwayne Johnson\\
    \Circled{9} Elon Musk &\Circled{10} Fei-Fei Li &\Circled{11} Geoffrey Hinton &\Circled{12} Jeff Bezos \\
    \Circled{13} Joe Biden &\Circled{14} Kamala Harris &\Circled{15} Marilyn Monroe &\Circled{16} Mark Zuckerberg \\
    \Circled{17} Michelle Obama &\Circled{18} Oprah Winfrey &\Circled{19} Renée Zellweger &\Circled{20} Scarlett Johansson \\
    \Circled{21} Taylor Swift &\Circled{22} Thomas Edison &\Circled{23} Vladimir Putin
    &\Circled{24} Woody Allen \\
    \Circled{25} Yann LeCun & \\
    \bottomrule
    \end{tabular}
    \label{tab:evaluate_ids}
\end{table}

\begin{table}[t]
    \centering
    \caption{\textbf{Quantitative comparison on style- and attribute-irrelevant prompts under single reference setting.} The best result is shown in \textbf{bold}, and the second best is \underline{underlined}. }
    \label{tab:quantitative_single_noattr}
    \resizebox{1\linewidth}{!}{
    \setlength{\tabcolsep}{1.5mm}
    \begin{tabular}{lccccc} 
        \toprule
         &  CLIP-T ($\uparrow$) &  CLIP-I ($\uparrow$) &  DINO ($\uparrow$) &  Face Sim. ($\uparrow$) &  Speed (s, $\downarrow$) \\ 
        \midrule
         Custom Diffusion~\cite{kumari2022multi} &  0.209&  0.751&  0.652 &  0.635&      548 \\ 
         FastComposer~\cite{xiao2023fastcomposer} &  0.210&  0.755&  0.624&  0.651&   3 \\ 
         IPAdapter~\cite{ye2023ip} &  0.212 &  0.752 &  0.649&  0.604&   3 \\ 
         PhotoMaker~\cite{li2023photomaker}  &  \textbf{0.231} &  0.710&  0.531 &  0.594 &  10 \\ 
         InstantID~\cite{wang2024instantid} &  0.205 &  0.758 &  \textbf{0.669} &  0.682 &  21 \\ 
         \midrule
         Ours &  \textbf{0.231}&  \underline{0.766}&  \underline{0.665} & \underline{0.686}&   4\\ 
         Ours w/o $\mathcal{L}_{edit}$ &  \underline{0.216} &  \textbf{0.768}&  0.662 & \textbf{0.688} &   4 \\ 
        \bottomrule
    \end{tabular}
    }
\end{table}

\begin{table}[t]
    \centering
    \caption{\textbf{Effect of layer selection for $\mathcal{L}_{edit}$.} }
    \label{tab:ablation_layer}
    \resizebox{0.9\linewidth}{!}{
    \setlength{\tabcolsep}{1.4mm}
    \begin{tabular}{ccccccc} 
        \toprule
         Upblock1 & Upblock2 & Upblock3 &  CLIP-T ($\uparrow$) &  CLIP-I ($\uparrow$) &  DINO ($\uparrow$) &  Face Sim. ($\uparrow$) \\ 
        \midrule
        \checkmark & &  &  0.232&  0.726 &  0.638 & 0.631 \\ 
        \checkmark & \checkmark &  &  0.233 &  0.718 &  0.632 &  0.608 \\ 
        \checkmark &  \checkmark &  \checkmark &  0.235 &  0.701 &  0.620&  0.568 \\ 
        \bottomrule
    \end{tabular}
    }
\end{table}

\begin{figure}[t]
   \begin{center}
   \includegraphics[width=1\linewidth]{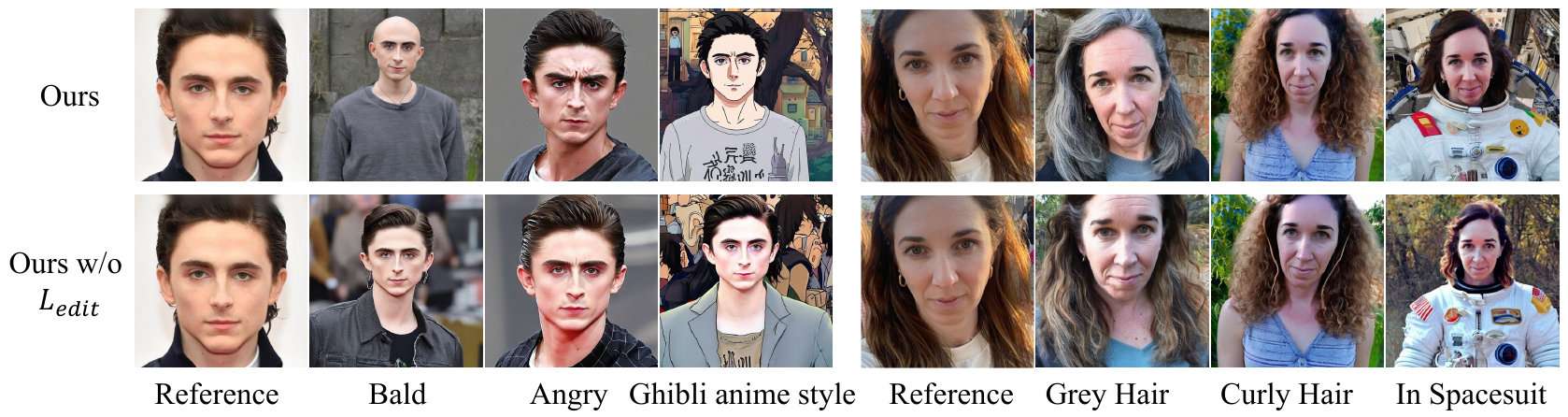}
   \end{center}
   \caption{\textbf{More ablation results on $\mathcal{L}_{edit}$.}} 
   \label{fig:ablation_editing}
\end{figure}

\begin{figure}[t]
   \begin{center}
   \includegraphics[width=.99\linewidth]{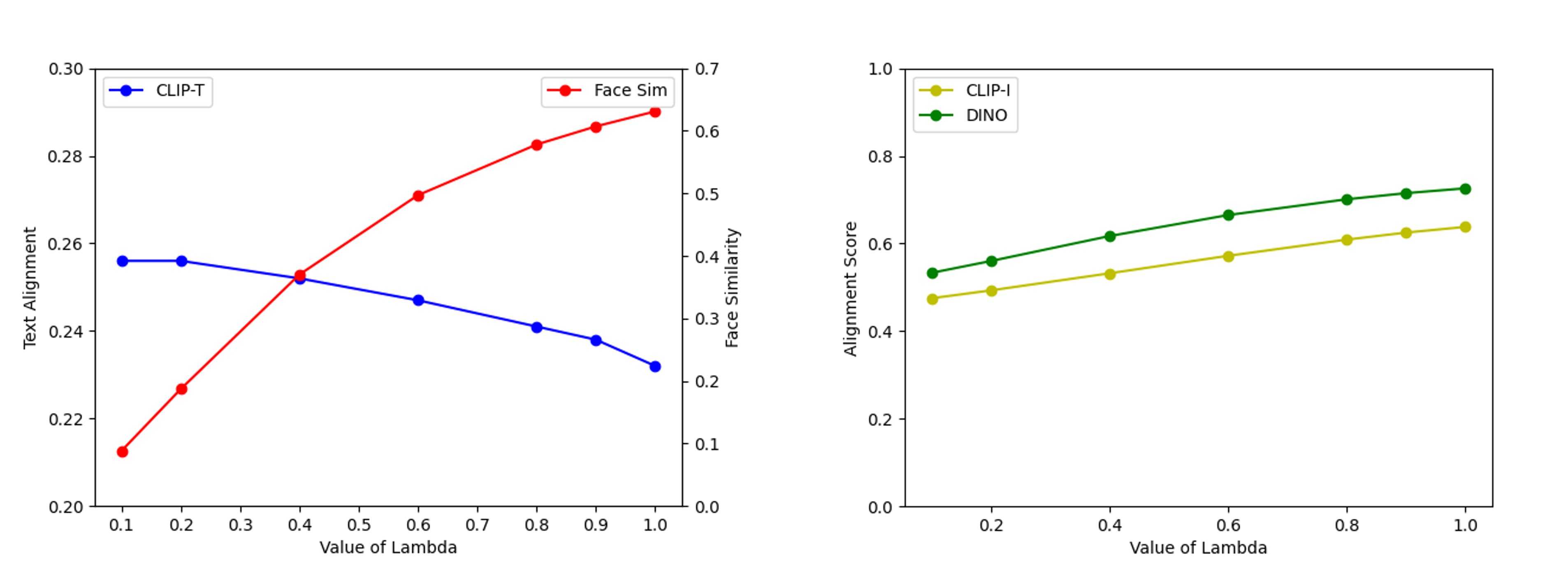}
   \end{center}
   \caption{\textbf{Ablation on the value of $\lambda$.} As $\lambda$ increases, the face similarity, CLIP-I and DINO improve, yet slightly decreases the text alignment (CLIP-T).} 
   \label{fig:ablation_lambda}
\end{figure}

\begin{figure}[t]
   \begin{center}
   \includegraphics[width=.99\linewidth]{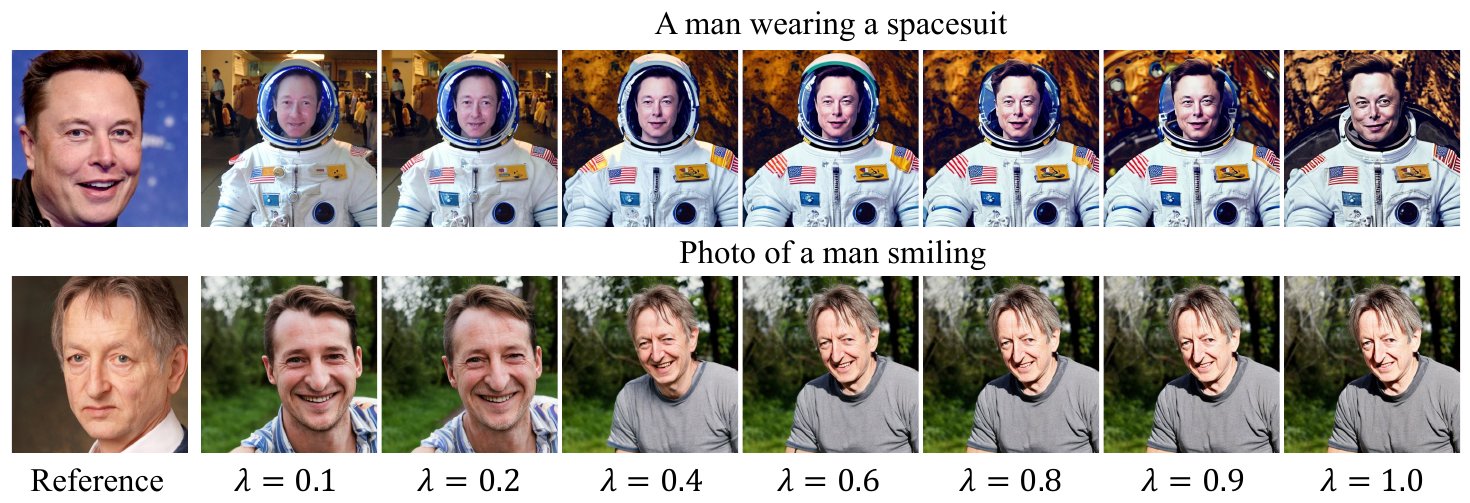}
   \end{center}
   \caption{\textbf{Visual comparisons by using different values of $\lambda$}.
   } 
   \label{fig:ablation_lambda_vis}
\end{figure}

\section{More Ablation Studies}
\label{sec:more_ablation}

\noindent \textbf{Effects of $\mathcal{L}_{edit}$ on metrics.}
As shown in Fig.~\ref{fig:ablation_editing},  $\mathcal{L}_{edit}$ could improve the editability (\eg, hair, expression, and style) while keeping the identity fidelity.
However, the CLIP-I and DINO metrics are calculated by comparing generated faces with reference faces. 
As for prompts editing style and attribute, the scores for successfully generated images can be lower than those of simply ID-preserved images.   
Table~\ref{tab:quantitative_single_noattr} further reports the metrics calculated on style- and attribute-irrelevant prompts.
One can see that our $\mathcal{L}_{edit}$ could improve text editability, while slightly affects the identity fidelity, demonstrating its effectiveness.

\subsubsection{Effect of layer selection for $\mathcal{L}_{edit}$.}

We have evaluated the impact of layer selection when calculating the editing direction loss $\mathcal{L}_{edit}$.
From Table~\ref{tab:ablation_layer}, incorporating Upblock2 and Upblock3 into the loss computation brings marginal improvements in editability at the cost of a significant decrease in identity consistency.
Therefore, we only utilize Upblock1 for loss calculation.

\subsubsection{Effect of $\lambda$.}

To evaluate the effect of hyper-parameter $\lambda$ during inference, we vary its value from 0.1 to 1, and the results are illustrated in Fig.~\ref{fig:ablation_lambda} and Fig.~\ref{fig:ablation_lambda_vis}.
It is observed that increasing $\lambda$ enhances identity fidelity.
Therefore, we keep $\lambda = 1$ during generation for better identity consistency.
However, users may choose to reduce $\lambda$ slightly to achieve better text alignment,  at the cost of a minor reduction in identity fidelity.

\subsubsection{Effect of the number of reference images.}

\begin{figure}[t]
   \begin{center}
   \includegraphics[width=1\linewidth]{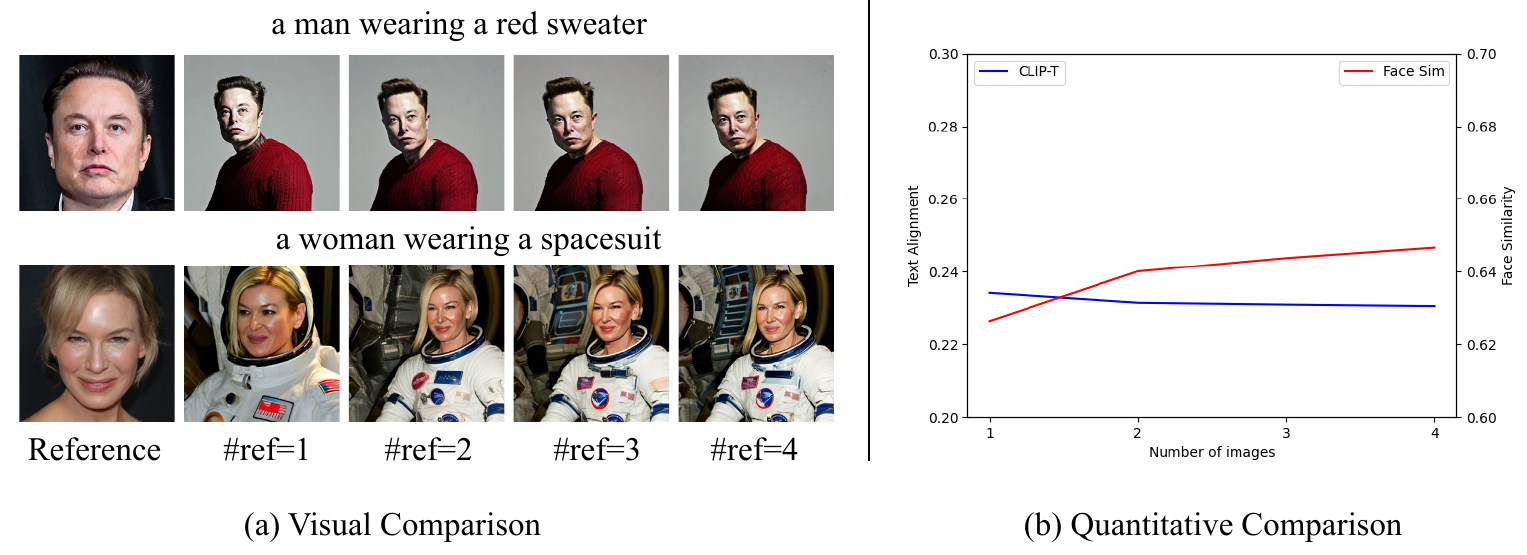}
   \end{center}
   \vspace{-2em}
   \caption{\textbf{Ablation study on the number of reference images.}
   As the number increases, the identity fidelity improves, yet slightly affecting the editability.
   } 
   \label{fig:ablation_number}
\end{figure}

We also examine the influence of the number of reference images.
As shown in Fig.~\ref{fig:ablation_number}, as the number of reference images increases, the identity fidelity of our MasterWeaver improves, yet slightly affects the editability.
Moreover, our method achieves superior identity fidelity and editability even with a single reference image.

\section{More Comparisons}
\label{sec:more_comparison}

\subsection{Comparison with InstantID}

\begin{figure}[t]
   \begin{center}
   \includegraphics[width=1\linewidth]{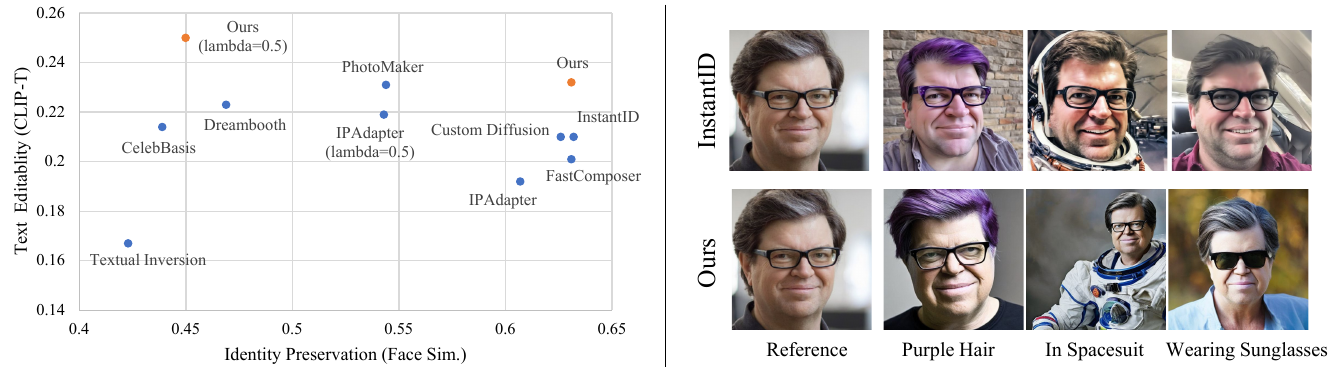}
   \end{center}
   \caption{\textbf{Comparison with Instant ID.}} 
   \label{fig:more_comparison}
\end{figure}

Fig.~\ref{fig:more_comparison} illustrates the comparisons between our method and InstantID~\cite{wang2024instantid}.
One can see that InstantID exhibits limited flexibility in facial poses and text editability.
In contrast, our method demonstrates better text alignment.
Fig.~\ref{fig:more_comparison} left also shows the metrics of identity preservation and text editability. Our method shows better trade-off between identity and editability.

\subsection{More Qualitative Results}

\begin{figure}[t]
   \begin{center}
   \includegraphics[width=.99\linewidth]{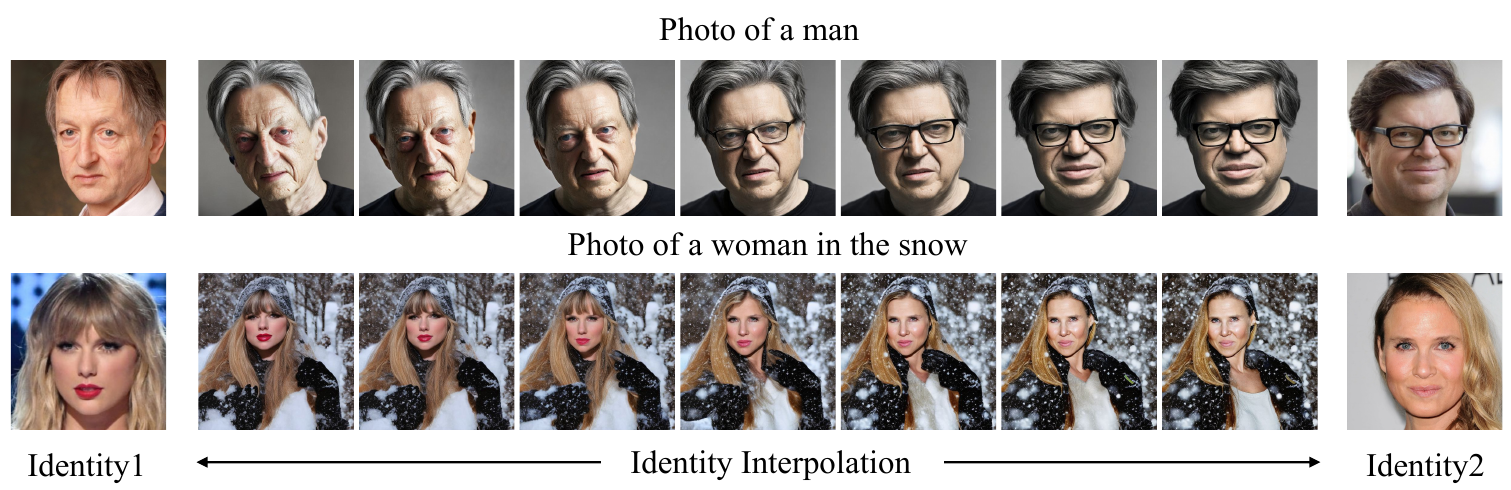}
   \end{center}
   \caption{\textbf{Interpolation between different identities.}
   } 
   \label{fig:interpolation}
\end{figure}

\begin{figure}[t]
   \begin{center}
   \includegraphics[width=.99\linewidth]{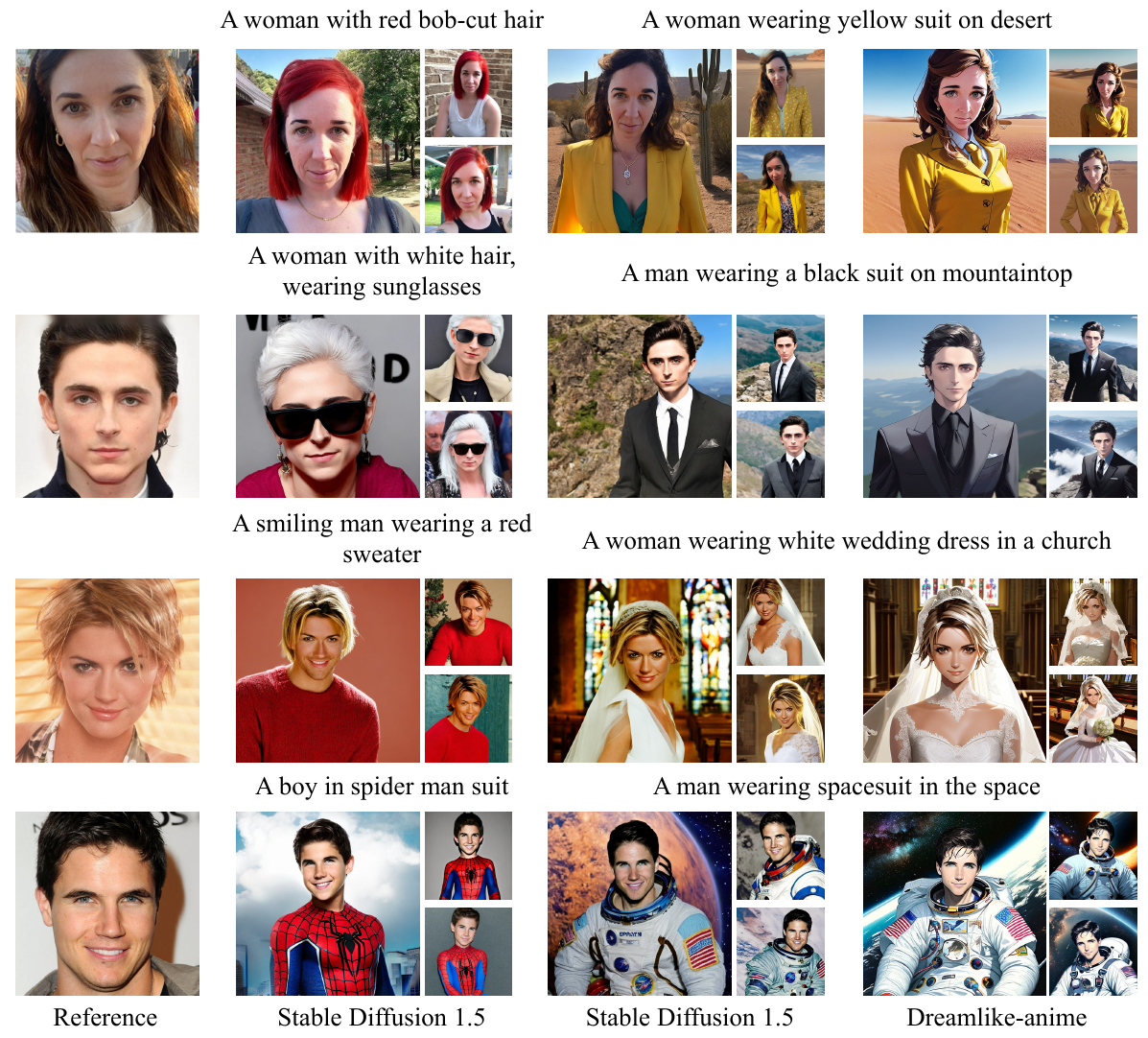}
   \end{center}
   \caption{\textbf{More visual results generated by our MasterWeaver.} 
   Our learned model can be directly combined with models fine-tuned from the stable diffusion 1.5, \eg, Dreamlike-anime.
   } 
   \label{fig:more_results}
\end{figure}

Figs.~\ref{fig:comparison_attrbute}$\sim$\ref{fig:comparison_bg} illustrate more comparisons between MasterWeaver and competing methods.
We see that MasterWeaver outperforms them, and generates photo-realistic images with diverse clothing, accessories, facial attributes, backgrounds, styles, and actions.
Fig.~\ref{fig:more_results} shows more results generated by our method.
As shown in the figure, our method can be directly combined with models fine-tuned from SD V1.5, \eg, Dreamlike-anime\footnote{https://huggingface.co/dreamlike-art/dreamlike-anime-1.0}.
Besides, with the obtained identity feature, our method can perform identity interpolation between different identities, as illustrated in Fig.~\ref{fig:interpolation}.

\subsection{More Quantitative Results}

\begin{figure}[t]
   \begin{center}
   \includegraphics[width=.99\linewidth]{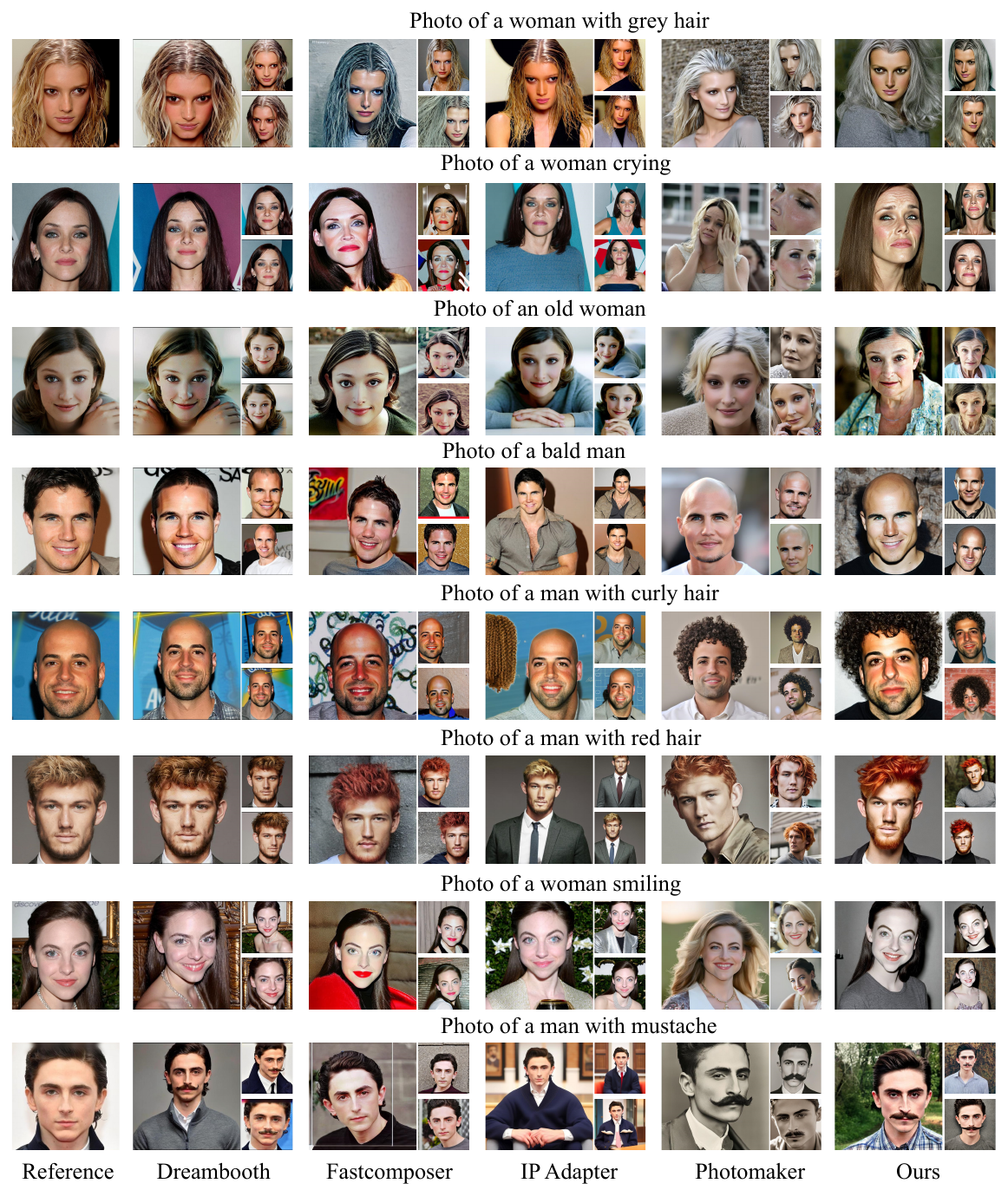}
   \end{center}
   \caption{\textbf{More visual comparisons for attribute editing.}
   } 
   \label{fig:comparison_attrbute}
\end{figure}

\begin{figure}[t]
   \begin{center}
   \includegraphics[width=.99\linewidth]{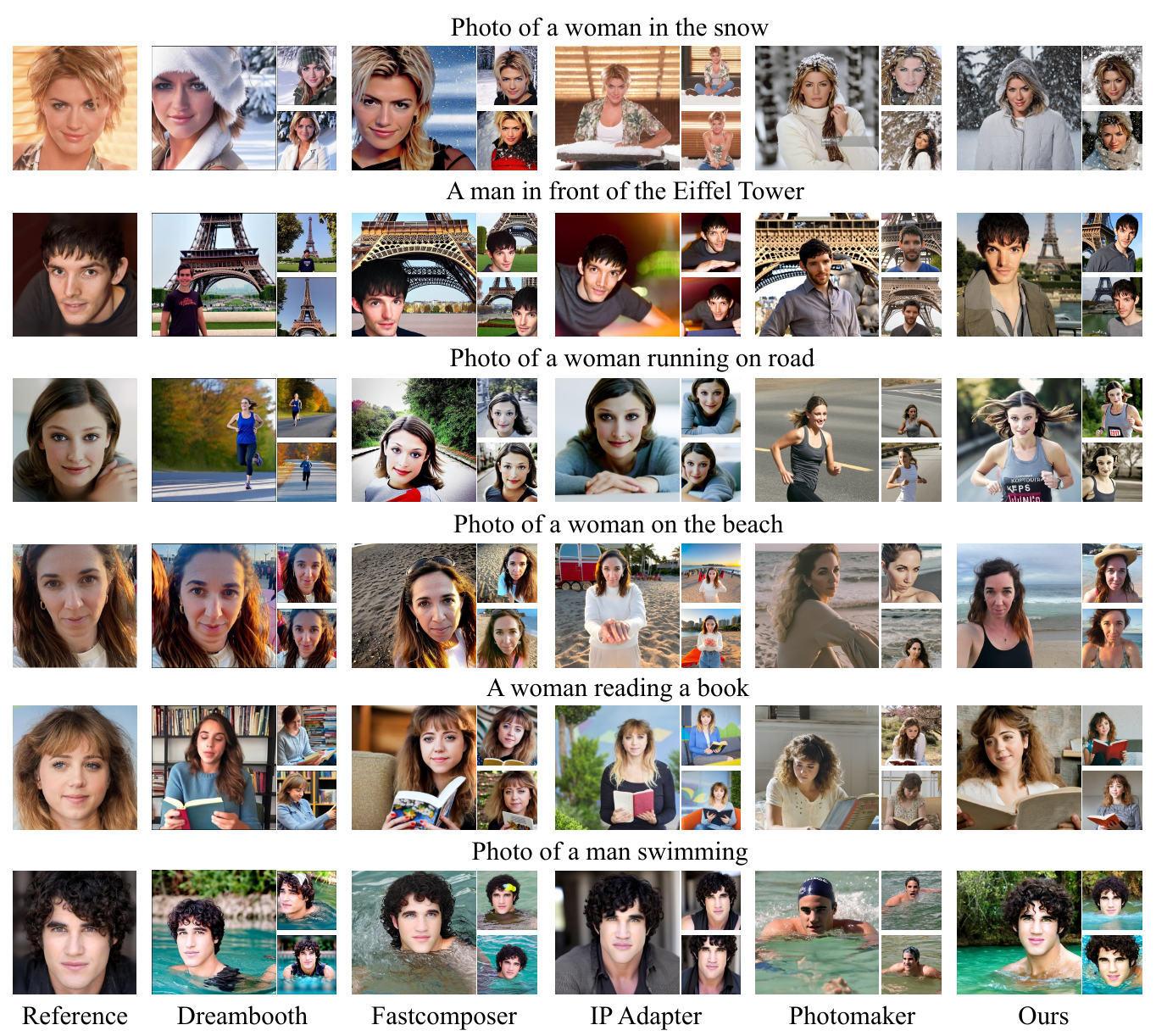}
   \end{center}
   \caption{\textbf{More visual comparisons for background and action generation.}
   } 
   \label{fig:comparison_bg}
\end{figure}

\noindent \textbf{Comparison on style- and attribute-irrelevant prompts.}
In our evaluation, we calculate the CLIP-I and DINO scores between the  faces extracted from generated images and reference images, and we employ different categories of prompts, including clothing, style, attribute, background, and action.
As shown in Figs.~\ref{fig:comparison_attrbute}$\sim$\ref{fig:comparison_bg}, our method generates images with flexible text editability.
As for prompts editing style and attribute, the methods with better editability may attain lower scores because the generated images can change the styles/attributes of reference images.  
In contrast, for methods with poor editability, \eg, FastComposer, the generated faces tend to copy the reference images, obtaining higher scores.
From Table~\ref{tab:quantitative_single_noattr}, when calculating metrics on style- and attribute-irrelevant prompts, our method exhibits better CLIP-I and comparable DINO metrics, showing its effectiveness.

\begin{table}[t]
\begin{center}
\caption{\textbf{User study.} The numbers indicate the percentage (\%) of volunteers who favor the results of our method over those of the competing methods.}
    \label{tab:user_study}
    \resizebox{1\linewidth}{!}{
    \setlength{\tabcolsep}{1.4mm}
    \renewcommand{\arraystretch}{1.2}
    \begin{tabular}{lccccc}
        \toprule
        Metric & \makecell[c]{Ours vs. \\ Dreambooth} & \makecell[c]{Ours vs. \\ CelebBasis}  & \makecell[c]{Ours vs. \\ Fastcomposer} & \makecell[c]{Ours vs. \\ IP Adapter} & \makecell[c]{Ours vs. \\ Photomaker} \\
        \midrule
        Text Fidelity & 68.8 & 61.5 & 62.2 & 70.2 & 52.1 \\
        Identity Fidelity & 64.0 & 70.1 & 55.3 & 53.5 & 66.8 \\
        Image Quality & 58.2 & 74.1 & 68.8 & 71.8 & 54.5 \\
        \bottomrule
    \end{tabular}
}
\end{center}
\end{table}

\noindent \textbf{User Study.} 
We invite volunteers to perform a user study on the personalized T2I results od MasterWeaver and its competitors. Given an identity image, a text prompt, and two synthesized images (ours \textit{v.s.} competitor's), the participants are asked to select the better one from three aspects. 
i) Text Fidelity: Which image is more faithful to the input text prompt?
ii) Identity Fidelity: Which image is more similar to the input person's identity? 
iii) Image Quality: Which generated image shows better quality? 

For each evaluation aspect, we employ 40 participants, and each participant is asked to evaluate 50 randomly selected sample images, \ie, 2000 responses in total.
As shown in Table~\ref{tab:user_study}, our method achieves a better trade-off between text editability and identity fidelity.
While our method receives similar preference to Photomaker regarding text fidelity, our identity fidelity is much better, demonstrating its superiority.

\begin{table}
    \centering
    \caption{\textbf{Text prompts used to construct the face-augmented dataset.}}
    \resizebox{0.8\linewidth}{!}{
    \setlength{\tabcolsep}{10mm}
    \begin{tabular}{c|c}
    \toprule
    Prompt  & Prompt \\
    \midrule
        face with mouth open  & face with mouth closed \\
        face with beard & face with eyeglasses \\
        smiling face & happy face  \\
        surprised face & angry face \\
        face with bangs & face with red hair \\
        face with black hair & face with blond hair \\
        face with grey hair & face with receding hairline \\
        face with curly hair & bald face \\
    \bottomrule
    \end{tabular}
    }
    \label{tab:augment_prompts}
\end{table}

\section{Limitation and Social Impact}
\label{sec:limitation}

\begin{figure}[t]
   \begin{center}
   \includegraphics[width=.99\linewidth]{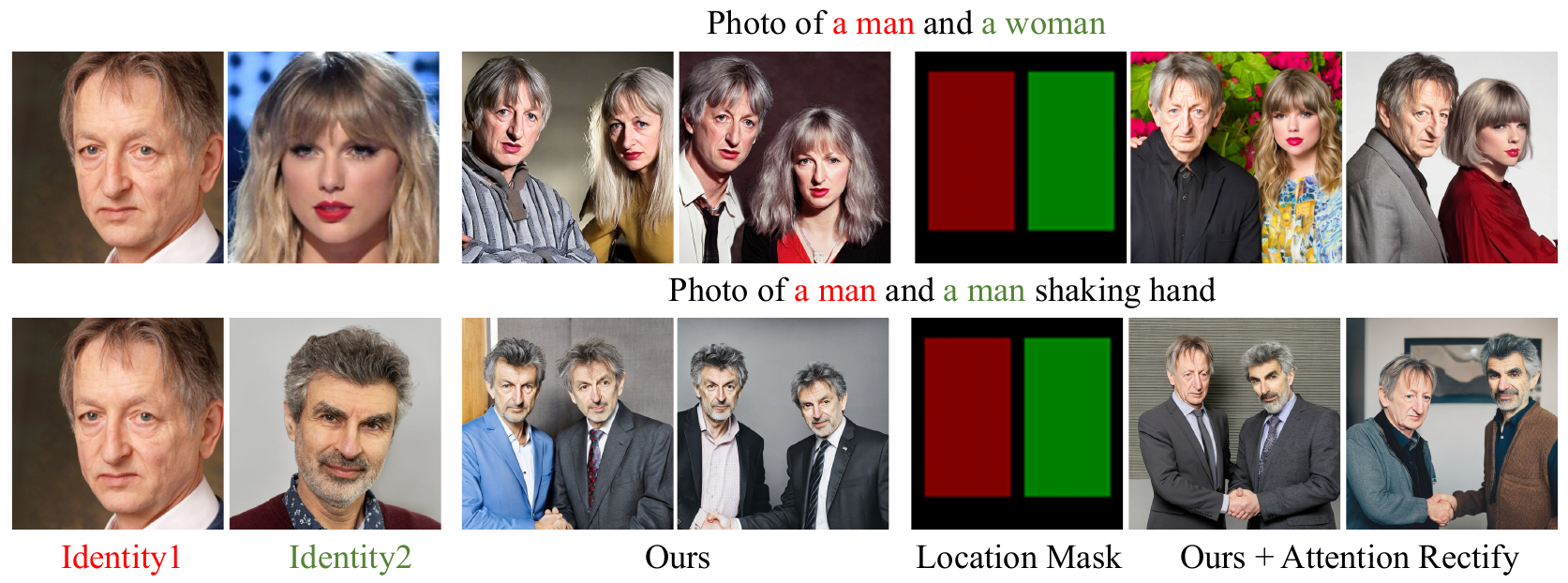}
   \end{center}
   \caption{\textbf{Failure cases of our MasterWeaver.}
   } 
   \label{fig:limitation}
\end{figure}

Our method is primarily trained to generate images with a single identity, and it may struggle with generating images consisting of multiple personalized identities. 
As illustrated in Fig.~\ref{fig:limitation}, directly combining two different identities fails to yield the intended results, instead producing a pair of faces that appear to be blends of the two.
Utilizing techniques, such as attention rectification, can mitigate this issue, but a pre-defined location map for each identity is needed.
In future work, we will explore more sophisticated approaches for the personalization of multiple identities.
Furthermore, our method's capability to generate photo-realistic images of specific individuals carries ethical considerations. 
Such a technique might be used for deepfake generation, particularly for offensive content. 
We will explore additional safeguarding methods to avoid this, such as integrating digital watermarks.

%
%

\begin{table}
    \centering
    \caption{\textbf{Text prompts used to calculate the editing direction loss.} The \texttt{<class word>} will be replaced with \texttt{woman}, \texttt{man}, and  \texttt{girl}, \etc.
    }
    \resizebox{1\linewidth}{!}{
    \setlength{\tabcolsep}{10mm}
    \begin{tabular}{c|p{0.8\linewidth}}
    \toprule
    Category  & Prompt \\
    \midrule
        \multirow{18}{*}{Source}  & a \texttt{<class word>} \\
        &a photo of a \texttt{<class word>} \\
        &a rendering of a \texttt{<class word>} \\
        &the photo of a \texttt{<class word>} \\
        &a photo of a clean \texttt{<class word>} \\
        &a photo of the cool \texttt{<class word>} \\
        &a bright photo of the \texttt{<class word>} \\
        &a cropped photo of a \texttt{<class word>} \\
        &a photo of the \texttt{<class word>} \\
        &a good photo of the \texttt{<class word>} \\
        &a photo of one \texttt{<class word>} \\
        &a rendition of the \texttt{<class word>} \\
        &a photo of the clean \texttt{<class word>} \\
        &a rendition of a \texttt{<class word>} \\
        &a photo of a nice \texttt{<class word>} \\
        &a good photo of a \texttt{<class word>} \\
        &a photo of the nice \texttt{<class word>} \\
        &a photo of a cool \texttt{<class word>} \\
    \hline
    \multirow{43}{*}{Target}   &  photo of a \texttt{<class word>} with straight bangs \\
        &photo of a \texttt{<class word>} with short bangs \\
        &photo of a \texttt{<class word>} with long bangs \\
        &photo of a \texttt{<class word>} with wavy hair \\
        &photo of a \texttt{<class word>} with curly hair \\
        &photo of a \texttt{<class word>} with straight hair \\
        &photo of a \texttt{<class word>} with short hair \\
        &photo of a \texttt{<class word>} with long hair \\
        &photo of a \texttt{<class word>} with black hair \\
        &photo of a \texttt{<class word>} with red hair \\
        &photo of a \texttt{<class word>} with purple hair \\
        &photo of a \texttt{<class word>} with yellow hair \\
        &photo of a \texttt{<class word>} with grey hair \\
        &photo of a \texttt{<class word>} with blue hair \\
        &photo of a \texttt{<class word>} with green hair \\
        &photo of a \texttt{<class word>} with blond hair \\
        &photo of a \texttt{<class word>} with rainbow hair \\
        &photo of a \texttt{<class word>} with hi-top\_fade hair \\
        &photo of a \texttt{<class word>} with bob-cut hair \\
        &photo of a \texttt{<class word>} with afro hair \\
        &photo of a smiling \texttt{<class word>} \\
        &photo of an angry \texttt{<class word>} \\
        &photo of a happy \texttt{<class word>} \\
        &photo of a chubby \texttt{<class word>} \\
        &photo of a sad \texttt{<class word>} \\
        &photo of a cute \texttt{<class word>} \\
        &photo of a bald \texttt{<class word>} \\
        &photo of a crying \texttt{<class word>} \\
        &photo of a surprised \texttt{<class word>} \\
        &photo of an old \texttt{<class word>} \\
        &photo of a young \texttt{<class word>} \\
        &photo of a \texttt{<class word>} with eyeglasses \\
        &photo of a \texttt{<class word>} with sunglasses \\
        &photo of a \texttt{<class word>} wearing eyeglasses \\
        &photo of a \texttt{<class word>} wearing sunglasses \\
        &photo of a \texttt{<class word>} wearing red hat \\
        &photo of a \texttt{<class word>} with lipstick \\
        &photo of a \texttt{<class word>} with arched eyebrows \\
        &photo of a \texttt{<class word>} with bushy eyebrows \\
        &photo of a \texttt{<class word>} with shallow eyebrows \\
        &photo of a \texttt{<class word>} with mustache \\
        &photo of a \texttt{<class word>} with goatee \\
    \bottomrule
    \end{tabular}
    }
    \label{tab:editing_prompts}
\end{table}

\begin{table}
    \centering
    \caption{\textbf{Text prompts used for quantitative evaluation.}
    The \texttt{<class word>} will be replaced with \texttt{man}, \texttt{woman}, \etc.
    }
    \resizebox{1\linewidth}{!}{
    \setlength{\tabcolsep}{10mm}
    \begin{tabular}{c|p{0.8\linewidth}}
    \toprule
    Category  & Prompt \\
    \midrule
    General  & a photo of a \texttt{<class word>} \\
    \hline
    \multirow{6}{*}{Clothing}   &  a \texttt{<class word>} wearing suit \\
       & a \texttt{<class word>} wearing a spacesuit \\
       & a \texttt{<class word>} wearing a red sweater \\
       & a \texttt{<class word>} in a chef outfit \\
       & a \texttt{<class word>} in a police outfit \\
       & a \texttt{<class word>} in Iron man suit\\
\hline
 \multirow{4}{*}{Accessory}&  a \texttt{<class word>} wearing a Christmas hat \\
        & a \texttt{<class word>} wearing a blue cap \\
        & a \texttt{<class word>} wearing sunglasses \\
        & a \texttt{<class word>} wearing a doctoral cap \\
\hline
 \multirow{9}{*}{Action}&  photo of a \texttt{<class word>} swimming \\
    & photo of a \texttt{<class word>} running on road \\
    & a \texttt{<class word>} is playing the basketball \\
    & a \texttt{<class word>} is playing the guitar \\
    & a \texttt{<class word>} plays the LEGO toys \\
    & a \texttt{<class word>} holding a bottle of red wine \\
    & a \texttt{<class word>} riding a horse \\
    & a \texttt{<class word>} walking the dog \\
    & a \texttt{<class word>} reading a book \\
    \midrule
 \multirow{17}{*}{Attribute}& photo of a chubby \texttt{<class word>} \\
         & photo of a young \texttt{<class word>} \\
         & photo of an old \texttt{<class word>} \\
         & photo of a baby \texttt{<class word>} \\
         & photo of an angry \texttt{<class word>} \\
         & photo of a surprised \texttt{<class word>} \\
         & photo of a \texttt{<class word>} crying \\
         & photo of a \texttt{<class word>} smiling \\
         & photo of a \texttt{<class word>} with mustache \\
         & photo of a bald \texttt{<class word>} \\
         & photo of a \texttt{<class word>} with long hair \\
         & photo of a \texttt{<class word>} with grey hair \\
         & photo of a \texttt{<class word>} with red hair \\
         & photo of a \texttt{<class word>} with blond hair \\
         & photo of a \texttt{<class word>} with purple hair \\
         & photo of a \texttt{<class word>} with short hair \\
         & photo of a \texttt{<class word>} with curly hair \\
         \hline
  \multirow{9}{*}{Style}& a \texttt{<class word>} Funko pop \\
        & a \texttt{<class word>} in Ghibli anime style \\
        & Manga drawing of a \texttt{<class word>} \\
        & a sketch of a \texttt{<class word>} \\
        & a \texttt{<class word>} in a comic book \\
        & a watercolor painting of a \texttt{<class word>} \\
        & a Greek marble sculpture of a \texttt{<class word>} \\
        & a black and white photograph of a \texttt{<class word>} \\
        & a pointillism painting of a \texttt{<class word>} \\
         \hline
     \multirow{4}{*}{Background} & photo of a \texttt{<class word>} in the snow \\
        & photo of a \texttt{<class word>} on the beach \\
        & photo of a \texttt{<class word>} on the sofa \\
        & a \texttt{<class word>} in front of the Eiffel Tower \\
        \bottomrule
    \end{tabular}
    }
    \label{tab:evaluate_prompts}
\end{table}

\begin{figure}[t]
   \begin{center}
   \includegraphics[width=.99\linewidth]{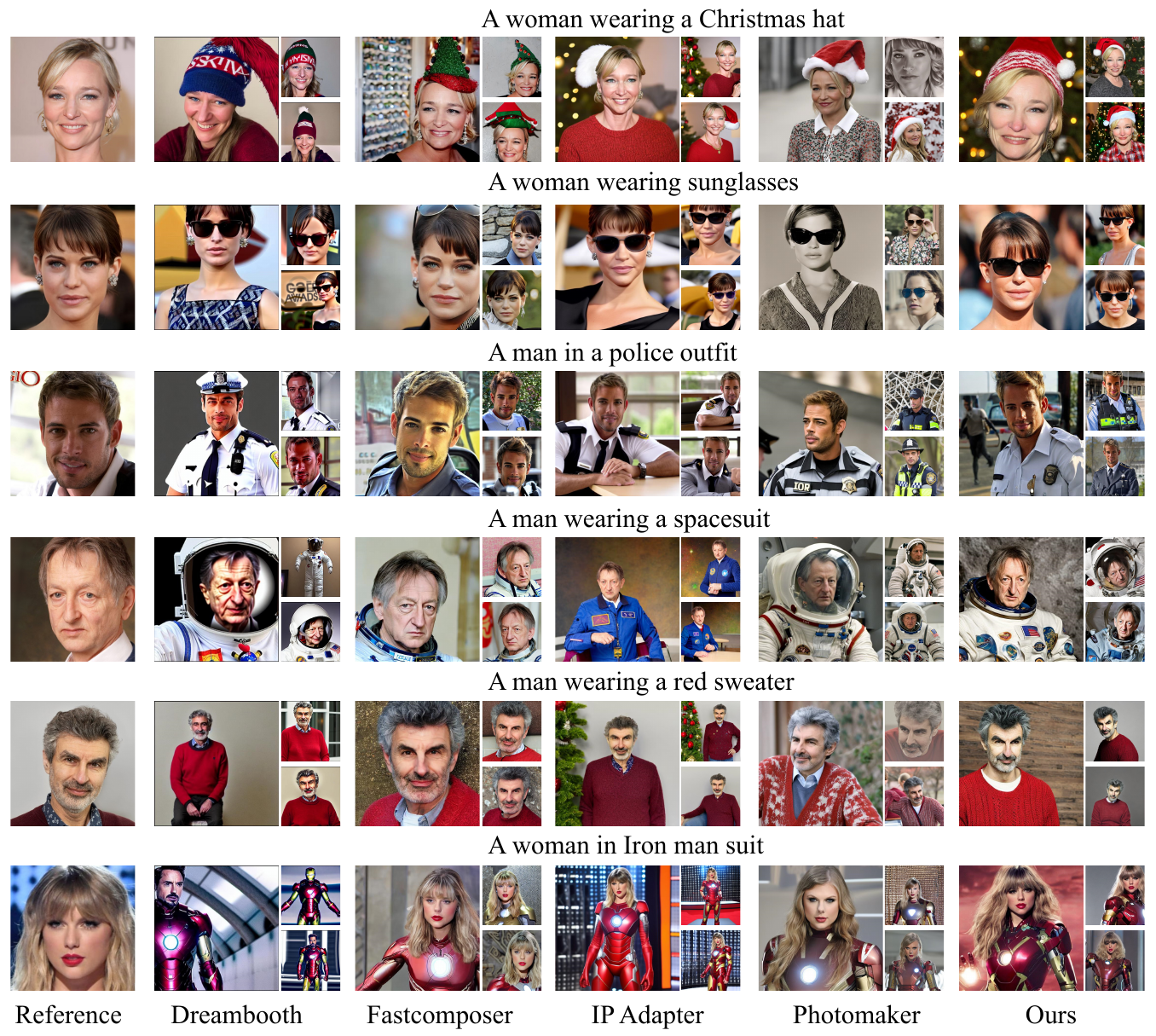}
   \end{center}
   \caption{\textbf{More visual comparisons for clothing and accessory generation.}
   } 
   \label{fig:comparison_clothing}
\end{figure}

\begin{figure}[t]
   \begin{center}
   \includegraphics[width=.99\linewidth]{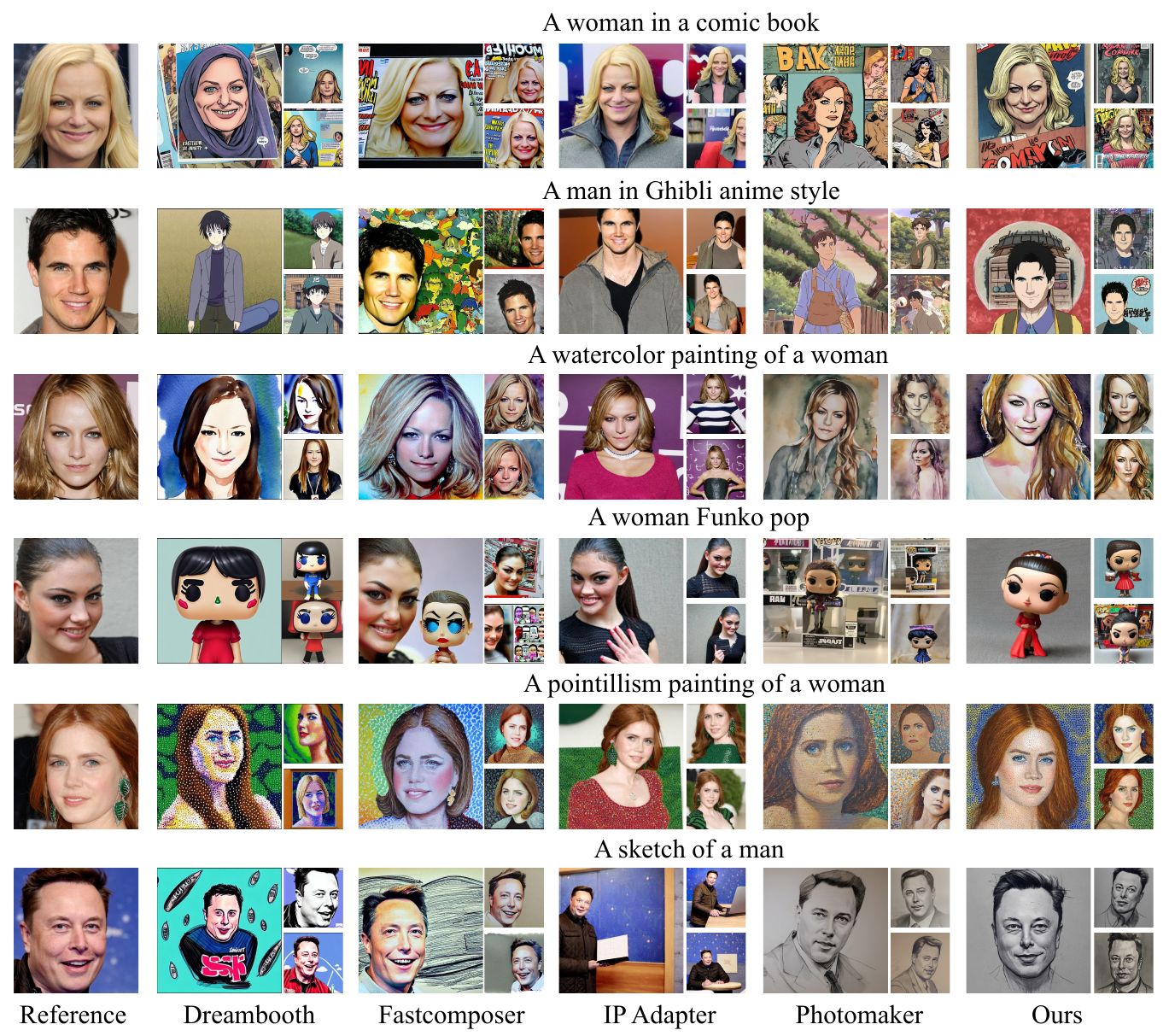}
   \end{center}
   \caption{\textbf{More visual comparisons for stylized image generation.}
   } 
   \label{fig:comparison_style}
\end{figure}

\end{document}